\documentclass[runningheads]{llncs}

\usepackage[T1]{fontenc}
\usepackage{graphicx}

\usepackage{orcidlink}
\usepackage[misc,geometry]{ifsym}
\usepackage{hyperref}
\usepackage{color}

\usepackage{algorithm}
\usepackage{algorithmic}
\usepackage{enumitem}
\usepackage{booktabs}

\usepackage{listings}
\usepackage{amssymb}
\usepackage{amsmath}

\usepackage{tikz}
\usetikzlibrary{backgrounds}
\usetikzlibrary{calc}

\usepackage{paper}
\usepackage{code}

\usepackage{footnote}
\makesavenoteenv{tabular*}
\makesavenoteenv{table*}

\DeclareMathOperator*{\argmax}{argmax}

\begin{document}

\author{Albert Nössig \inst{1,2} (\Letter) \orcidlink{0000-0001-7688-9836} \and
Tobias Hell \inst{2} \orcidlink{0000-0002-2841-3670} \and
Georg Moser \inst{1} \orcidlink{0000-0001-9240-6128}}

\authorrunning{A. Nössig et al.}

\institute{Department of Computer Science, University of Innsbruck,
Tyrol, Austria\\
\email{georg.moser@uibk.ac.at} \and
Data Lab Hell GmbH, Europastraße 2a, 6170 Zirl, Tyrol,
Austria\\
\email{\{albert.noessig, tobias.hell\}@datalabhell.at}\\ 
}
\title{A Voting Approach for Explainable Classification with Rule Learning}

\maketitle

\begin{abstract}
State-of-the-art results in typical classification tasks are mostly
achieved by unexplainable machine learning methods, like deep neural
networks, for instance. Contrarily, in this paper, we investigate the
application of rule learning methods in such a context. Thus,
classifications become based on comprehensible (first-order) rules,
explaining the predictions made. In general, however, rule-based
classifications are less accurate than state-of-the-art results (often significantly).
As main contribution, we introduce a voting
approach combining both worlds, aiming to achieve comparable results
as (unexplainable) state-of-the-art methods, while still providing
explanations in the form of deterministic rules.
Considering a variety of benchmark data sets including a use case of
significant interest to insurance industries, we prove that our
approach not only clearly outperforms ordinary rule learning methods,
but also yields results on a par with state-of-the-art outcomes.

\end{abstract}

\section{Introduction}
\label{Introduction}

There has been an immense progress in the field of machine learning in the last years. In particular, deep learning has led to the development of very complex models which yield highly accurate results. Especially the extensive use of large transformer models for text (eg.~\cite{NEURIPS2020_1457c0d6}) and image to text generation (eg.~\cite{rombach2021highresolution}) has led to various applications perceived as milestones in artificial intelligence by parts of society as they become more and more accessible. However, these approaches lack explainability and interpretability. To overcome this defect, the field of \emph{Explainable Artificial Intelligence} (\emph{XAI} for short, see for example~\cite{xai,xai2,Storey:2022}) has recently significantly increased momentum.

Indeed, in many application areas of machine learning, like automotive, medicine, health and insurance industries, etc.,\ the need for security and transparency of the applied methods is not only preferred but increasingly often of utmost importance. Arguably, the lack of progress in self-driving cars can be attributed to unintuitive decisions in autonomous vehicles in edge cases, that is,
situations that have not been foreseen in training. In such situations safety of the behaviour of autonomous vehicles cannot currently be guaranteed, cf.~\cite{Kirkpatrick:2022}.

A textbook example of XAI is the generation of deterministic rules which can be used for classification. 
Thinking for instance of predicting the health status of a patient, both the doctor as well as the patient will have more trust in a prediction given by a rule which they can understand as shown in the example below instead of a prediction which exaggeratedly
said seems to come out of nowhere when, for instance, a neural network is consulted.

\begin{lstlisting}[style=Prolog]
		IF	BloodPressure in [70,80]
		  AND	Insulin in [140,170]
		THEN	Diabetes = Yes.

\end{lstlisting} 

The resulting explanations for the made predictions in the form of causal rules make such approaches especially desirable since they can be categorized as \emph{most informative} in the area of XAI~\cite{hulsen}. A classical example in this context is the
field of \emph{Rule Induction}~\cite{F12}, which investigates the construction of simple if-then-else rules from given input/output examples in order to solve a given (classification) problem (cf. Section~\ref{RuleLearning}). Representative examples of such rules are shown in Section~\ref{Approach}, illustrating the major advantages of rule learning methods, namely their transparency and comprehensibility, which make them a desirable classification tool in many areas. Unfortunately, these benefits come with the drawback of generally less accurate results (cf. Table~\ref{accuracies}). Moreover, for a long time it has not been possible to efficiently apply rule learning methods on vast data sets~(cf.~\cite{mitra18}) as for instance present in the case study considered in Section~\ref{CaseStudy}. 
In previous work~\cite{modularapproach} we have extensively investigated this issue and as a solution we introduced a modular approach (cf. Section~\ref{RelatedWork}) which enables the application of ordinary rule learning methods such as, for instance, \foil~\cite{foil} and \ripper~\cite{ripper} (cf. Section~\ref{Techniques}) on very large data sets including several hundreds of thousands examples. 
However, the in general poorer performance than state-of-the-art methods wrt. accuracy remains.

\begin{figure}[t]
\caption{\textbf{Motivational Example:} Procedure of the proposed voting approach with the example of the MNIST digits distinguishing between the two basic scenarios, namely coinciding predictions given by the rule learners as well as conflicting ones. In a first step only the explainable methods are considered using the corresponding prediction in case they match. Otherwise, an (unexplainable) state-of-the-art method -- the so-called \emph{decider} -- is consulted to resolve the existing rule conflict.}
\label{fig:votingApproach}  
\begin{center}
  \setlength{\fboxsep}{0pt}
  \fbox{
    \includegraphics[width = \textwidth]{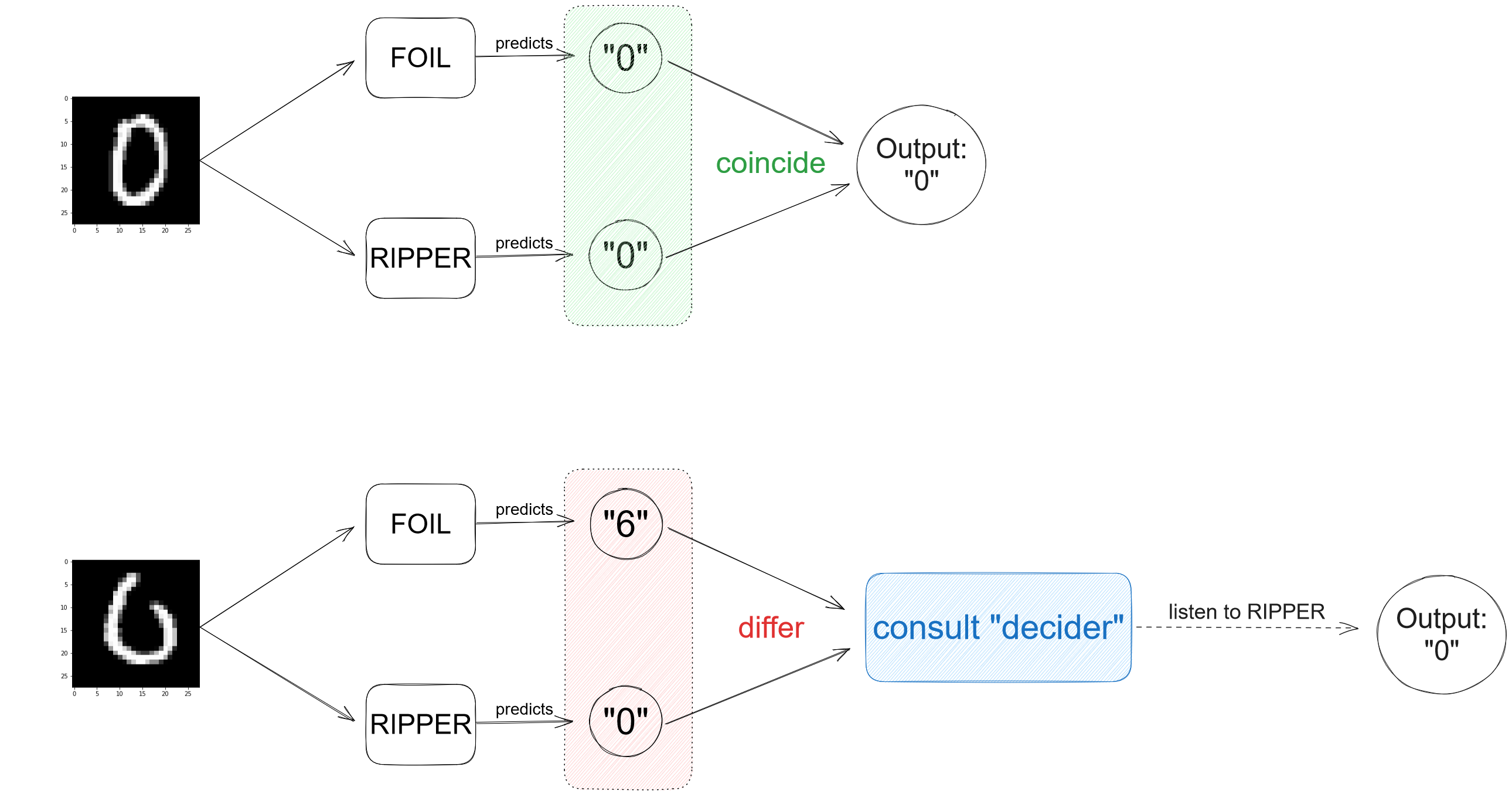}
  }
\end{center}
\end{figure}

In this paper, we build upon our previous work exploiting the possibility to apply rule learning methods on very large data sets and introduce a novel voting approach combining these methods with the state-of-the-art predictions from unexplainable ML tools to eliminate the disadvantage of worse outcomes, while preserving the interpretability of the obtained results. Concerning this, we need to balance between full explainability and best-possible accuracy.
In consultation with our collaboration partner from insurance business -- \emph{Allianz Private Krankenversicherung (APKV)} -- it is even more important to ease the understanding of a classification made than to make the whole procedure fully transparent. Therefore, the approach introduced in this paper involves beside the explainable methods also a black-box method which is consulted in case of distinct predictions given by the rule learners. In this case the unexplainable method serves as \emph{decider} solving the present rule conflict by suggesting us one rule learner on which we should listen as discussed in more detail in Section~\ref{Approach} and illustrated in the motivational example shown in Figure~\ref{fig:votingApproach}. 
This procedure goes hand in hand with our understanding of end-to-end explainability because similar to many other methods for rule conflict resolution (cf. for instance~\cite{Lindgren}) the unexplainable method suggests the appropriate rule to use for classification but the prediction itself, nevertheless, remains comprehensible for human beings.

To implement this project and for experimental evaluation we consider in particular the rule learning methods \foil\ and \ripper\ as well as decision trees as explained in more detail in Section~\ref{RuleLearning}. Moreover, to illustrate the versatility of the applicable methods, we additionally considered \emph{CN2}~\cite{CN2} as another explainable method and applied the corresponding python implementation from the \emph{Orange} package in our experiments. As unexplainable methods we apply neural networks or boosted decision trees as shown in Section~\ref{Experiments} where we demonstrate that our proposed voting approach yields accuracies on a par with highly complex unexplainable ML methods. Apart from evaluating our approach on common data sets from the \emph{UCI Machine Learning Repository}~\cite{Dua:2019} as well as the MNIST images~\cite{lecun2010mnist,fmnist}, we also present novel results
on explainable \emph{classifications of dental bills}. The latter case study stems from an industrial collaboration with  \emph{Allianz Private Krankenversicherung (APKV)} which is an insurance company offering diverse services in Germany.

Summed up, our main goal is to solve a classification problem by applying easily comprehensible rules and achieve accuracies comparable to state-of-the-art outcomes. At the end of the day we especially aim for \emph{end-to-end explainablity} in order to better understand the underlying data (cf. Section~\ref{DetailedAnalysis}). With the example of insurance industries, the proposed approach and the eventually obtained rules with high accuracy can be used in practice to systematically find inconsistencies in the training data which can then later be discussed with experts in the field in order to improve data quality. All in all, this paper directly builds on our previous work and expands upon the approach presented therein to not only solve problems with large amounts of data using explainable methods, but also to achieve satisfactory results which are on a par with the outcomes of unexplainable state-of-the-art methods.   

More precisely, we make the following contributions.
\subsubsection{Voting Approach Based on Rule Learning.} We introduce a novel voting approach based on rule learning combining the benefits from explainable and unexplainable methods by applying an ensemble of both transparent as well as black-box models
  (see Section~\ref{Approach} for further details.) Together with the modular approach introduced in our previous work~\cite{modularapproach}, this makes rule learners a serious alternative to state-of-the-art classification tools.

\subsubsection{Experimental Evaluation.} Further, we provide ample experimental evidence that our methodology provides significant
  improvements on the accuracy of common rule learning methods for the standard benchmarks
  (see Section~\ref{Experiments}).

\subsubsection{Industrial Use Case.} Finally, we show that our approach yields accuracies close to stochastical, non-explainable classification methods for the case study, while giving reasons for the resulting predictions in the form of deterministic rules which are of direct interest to our industrial collaboration partner
  (see Section~\ref{CaseStudy}).

\subsubsection{Overview.}

In Section~\ref{Notations} we introduce the major definitions and notations as well as the general ideas behind the methods on which our voting approach is based on. 
Section~\ref{RelatedWork} serves to discuss related work focusing on similar goals as considered in this paper, especially on various forms of explainable classification, while we concretely introduce our aforementioned voting approach in Section~\ref{Approach}. Section~\ref{Experiments} provides ample evidence of the advantages of our approach
and presents the case study mentioned. Finally, in Section~\ref{Conclusion} we summarise the main results and discuss ideas for future work. To keep the presentation succinct, we have delegated the description and comparison
of the employed rule learning techniques, namely~\foil\ and \ripper, to the Appendix, see
Section~\ref{Techniques}.

\section{Notations \& Preliminaries}
\label{Notations}

In order to enhance the performance of interpretable methods we apply ensemble learning in this paper, which we briefly describe in the following.

\subsection{Ensemble Learning}

Ensemble learning (see for instance \cite{10.5555/2381019}) is an often used approach in machine learning, where several models are combined to increase the accuracy of the predictions. The most common ensemble methods are summarised in the following. 

\begin{enumerate}
\item \textbf{Voting.} This is one of the simplest ensemble methods used for classification. First, various models are trained and a prediction is given by each model, while the final prediction is the one with the most votes. This approach can be extended by assigning weights to the individual models. In classification problems it is also common to apply \emph{soft voting} where the models output possibilities of belonging to a certain class and the final output is the (weighted) average of possibilities.   
\item \textbf{Bagging.} In \emph{Bootstrap AGGregatING} multiple models are generated by using the same algorithm with subsets of the original data set randomly created by applying \emph{bootstrap sampling}~\cite{EfroTibs93}. The outputs of the base learners are then aggregated by voting.
\item \textbf{Stacking.} Multiple so-called \emph{first-level} models are generated in a first step using the original input data. Afterwards, an additional \emph{Meta-Learner} is trained using the outputs of the first-level models as input features while the original labels are still regarded as labels. This \emph{second-level} learner aims to find the best possible way to combine the first-level learners. 
\item \textbf{Boosting.} An ensemble is built sequentially by training a model with the same data set but the instances are weighted according to the error of the last prediction with the main idea of forcing the model to focus on instances that are hard to predict.\\
\end{enumerate}

Instead of several unexplainable methods, the ensemble approach proposed in this paper combines a single unexplainable state-of-the-art method with explainable rule learning tools such as defined in the following.

\subsection{Rule Learning}
\label{RuleLearning}

As already mentioned in the introduction, \emph{Rule Induction} focuses on learning simple if-then-else rules such as the following example from Fürnkranz et al.~\cite{F12}. A data set containing the variables \emph{Education, Marital Status, Sex, Has Children} and \emph{Car} is given and the aim is to learn rules concerning the first four variables to predict the type of car. For instance, the following rule is learned:

\begin{lstlisting}[style=Prolog]
		IF	Sex = Male
		  AND	Has Children = Yes
		THEN	Car = Family.

\end{lstlisting}

We are in particular interested in the method \ripper~\cite{ripper} which is state-of-the-art in this field. Its basic procedure is explained in detail in the Appendix in Section~\ref{Techniques}.

A related topic to Rule Induction is \emph{Inductive Logic Programming} (\emph{ILP}), which investigates  the  construction  of  first- or higher-order  logic  programs.
In the context of this paper, it suffices to conceive the learnt hypothesis as first-order Prolog clauses as depicted below.
\begin{lstlisting}[style=Prolog]
  H :- L1, ..., Lm
\end{lstlisting}
Here, the \emph{head} $H$ is an atom and the \emph{body} $L_1, \dots, L_m$ consists of literals, that is atoms or negated atoms. An extensive overview on ILP is given by Cropper et al.~\cite{ilp-comp}.  

Concerning ILP, especially one of the first tools from this field, the \foil\ algorithm \cite{foil}, is of main interest for us due to its simplicity. Note that our \emph{Python} reimplementation (cf.~Section~\ref{Techniques}) is able to handle large data sets straight away which is in particular beneficial for the data set considered in our case study. Moreover, contrarily to more modern ILP-tools which rather aim to generate complex (recursive) programs, \foil\ is very well suited to learning simple if-then-else rules as we want to generate.

\begin{figure}[t]
\caption{Visualisation of a decision tree constructed for the nominal weather data set.}
\label{fig:tree}
\smallskip
\begin{center}
  \setlength{\fboxsep}{0pt}
  \fbox{
    \includegraphics[width = 0.95\textwidth]{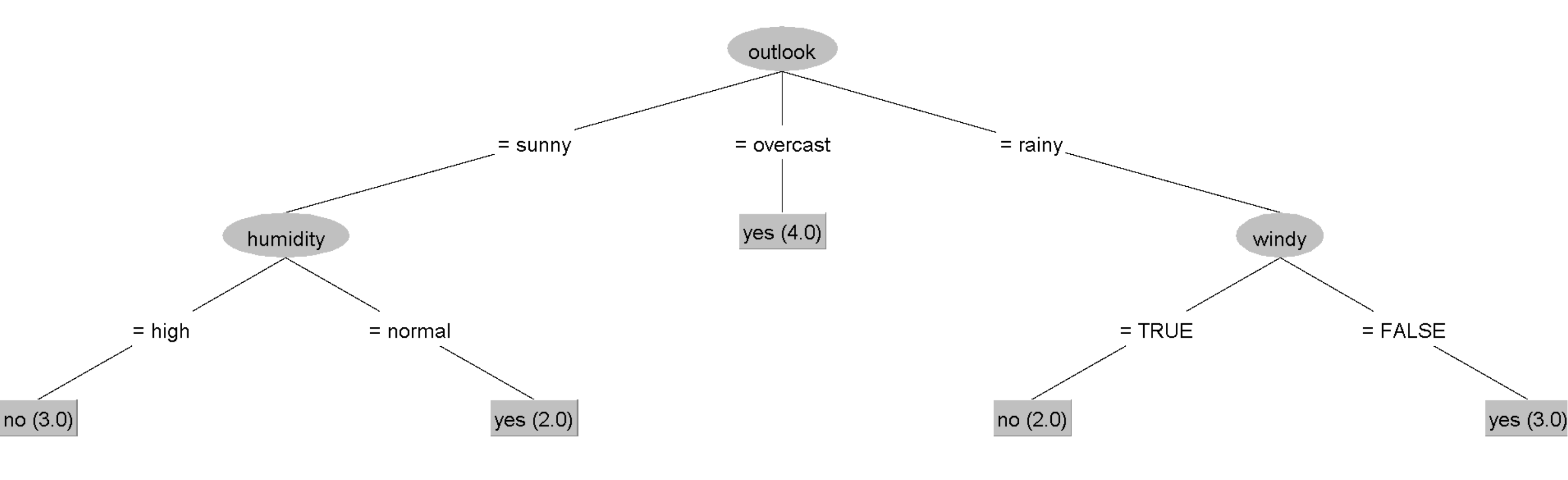}
  }
\end{center}
\end{figure}

Finally, the outcomes produced by decision trees (see \cite{trees} for instance) can be interpreted as rules. Consider for instance the tree constructed for the nominal weather data set \footnote{See \url{https://www.kaggle.com/datasets/nicolausboby/weathernominal}.} shown in Figure~\ref{fig:tree} where the leftmost leaf corresponds to the following if-then-else rule:

\begin{lstlisting}[style=Prolog]
		IF	Outlook = Sunny
		  AND	Humidity = High
		THEN	Play = Yes.

\end{lstlisting}

So, we also consider decision trees as rule learning methods in the following although we use them mostly for comparison with the above-mentioned methods because trees are not reasonably applicable within our voting approach. 
Shortly put, the kind of rules obtained from a decision tree is not 
suitable for our purposes.
A more detailed explanation can be found
in Section~\ref{Experiments}.

\section{Related Work}
\label{RelatedWork}

Especially in fields where decisions can deeply affect human lives attempts are being made to replace pure black-box models with interpretable methods and in some cases it has already been shown that explainability does not always go hand in hand with a loss of accuracy~(cf. \cite{Rudin2019Why}). However, in general, explainable methods are still outperformed by complex models which are not understandable for humans (cf. for instance~\cite{gulum,angelov}). In order to circumvent this disadvantage, in this paper we make use of ensemble learning with the aim to enhance the performance of interpretable methods.

In existing literature~\cite{s22197268,iasc.2023.032262,10.3389/fpubh.2022.874455}, two commonly used explainability approaches are LIME~\cite{lime} and SHAP~\cite{shap}. Both are local surrogate models yielding an approximated contribution of each input for a specific prediction in a local neighbourhood, see for instance \cite{surveyXAI2021_Burkart_Huber}. Such local approaches are only valid for the considered sample and cannot be generalised to the entire machine learning model.
Moreover, they are post-hoc approaches rather giving insights in the prediction made by a neural network, for instance, by highlighting which features of the input have been decisive for the obtained result. A common example in this context~\cite{lime} is a neural network that has been trained to distinguish between a husky and a wolf yielding high accuracies on the training data but miserably failing on examples that have not been used during training of the network. LIME has been applied in this context showing that the prediction of the neural network was based on whether there was snow in the given image because all images of wolves used during training had snow in the background. Similarly, in fields where in particular personal informations of humans are considered LIME might be helpful for example to reveal that the applied model is biased. So, these methods can be used for instance to identify problems regarding neural network training but do not directly yield information about the data per se. Note that \emph{Anchors}~\cite{anchors} are a successor of LIME improving especially the level of interpretability of the results but still applied post-hoc on a given machine learning tool rather explaining the decision making of the black-box method. In this context Elton~\cite{elton} states that "deep neural networks are generally non-interpretable" and evaluates the above-mentioned methods together with similar post-hoc approaches.

As opposed to the above-mentioned methods, we aim to achieve explainability of the obtained results by corroborating the made predictions with comprehensible rules as shown in more detail in Section~\ref{Approach}. Beside the application of the rules for classification they can also be used to gain a general insight in the underlying data and reveal inconsistencies (cf. \cite{modularapproach} and Section~\ref{DetailedAnalysis}).\\

Our proposed voting approach is principally based on the modular methodology introduced in our previous work~\cite{modularapproach}, where we split the whole procedure from given input examples to generating rules describing them into three independent phases (cf. Figure~\ref{fig:modularapproach}). First of all, we aim to reduce the complexity of the given input data during the \emph{Representation Learning} phase where common dimensionality reduction methods are involved. The compact form of the input examples is advantageous for the second phase where clustering is applied aiming for a sophisticated \emph{Input Selection} where the original problem is divided into several independent smaller problems on which the rule learner is applicable without any restrictions. Nevertheless, note that the rule learner is applied on the data in its original form in order to preserve explainability. By dividing the original problem into several subproblems with similar positive examples and heterogeneous negative examples as input we not only enable the application of rule learning methods on very large data sets but also the time consumption for rule generation can be reduced by a multiple because the rules on each subproblem can be learned in a parallel manner. Afterwards, the rule sets obtained on each subproblem are concatenated and can be used to classify the original problem without a significant loss of accuracy.
Summed up, this way of proceeding makes it possible to apply rule learning methods on very large data sets as for instance the MNIST digits.

\tikzstyle{txtlabel} = []
\tikzstyle{method} = [shape=rectangle, rounded corners, draw, fill=white, minimum width=2.2cm, minimum height=1.2cm]
\begin{figure}[t]
\caption{Modular Approach to Rule Learning. The first phase (\emph{Representation Learning}) is intended to yield a compact representation of the original (high-dimensional) input data. This is advantageous for clustering applied subsequently during the second phase (\emph{Input Selection}). These two steps put in front of the application of a chosen \emph{Rule Learner} in the final phase make it possible to find comprehensible rules on very large data sets in reasonable time.}
\label{fig:modularapproach}
\smallskip
\centering
\begin{minipage}{\linewidth}
  \begin{center}
  \begin{tikzpicture}[scale=1]
  \begin{scope}
    \node[method] (represent) at (0,0) {
      \begin{minipage}{2.6cm}
        \textbf{\small Representation Learning}
      \end{minipage}};    
    \ibox{(0,-1.5)} {2.2}{1}{\strut{}...};
    \ibox{(0.1,-1.9)}  {2.2}{1}{\tiny Neural Networks};
    \ibox{(0.2,-2.3)}  {2.2}{1}{\tiny UMAP};
  \end{scope}
  \begin{scope}
  \node[method] (clustering) at (4cm,0) {
    \begin{minipage}{2.7cm}
      \textbf{\small Input Selection}
    \end{minipage}
  };
    \ibox{(4,-1.5)} {2.2}{1}{\strut{}...};
    \ibox{(4.1,-1.9)}  {2.2}{1}{\tiny $k$-means};
    \ibox{(4.2,-2.3)}  {2.2}{1}{\tiny DBSCAN};  
  \end{scope}
  \begin{scope}
  \node[method] (learner) at (8cm,0) {
    \begin{minipage}{2.6cm}
      \textbf{\small Rule Learner}
    \end{minipage}
  };
   \ibox{(8,-1.5)} {2.2}{1}{\strut{}...};
    \ibox{(8.1,-1.9)}  {2.2}{1}{\tiny \foil};
    \ibox{(8.2,-2.3)}  {2.2}{1}{\tiny \ripper};
  \end{scope}
  \node[minimum width=1.3cm,anchor=west,align=left] (answer) at ($(learner.east)+(6mm,0)$) {    
\begin{lstlisting}[style=prolog,basicstyle=\color{darkviolet}\scriptsize\ttfamily,frame=none]
target(V) :-
  black_1(V),
  black_N(V).
\end{lstlisting}
  };
  \begin{scope}[on background layer]
    \draw[->] (represent) -- (clustering);
    \draw[->] (clustering) -- (learner);
    \draw[->] (learner) -- (answer);
  \end{scope}
\end{tikzpicture}
\end{center}
\end{minipage}
\end{figure}
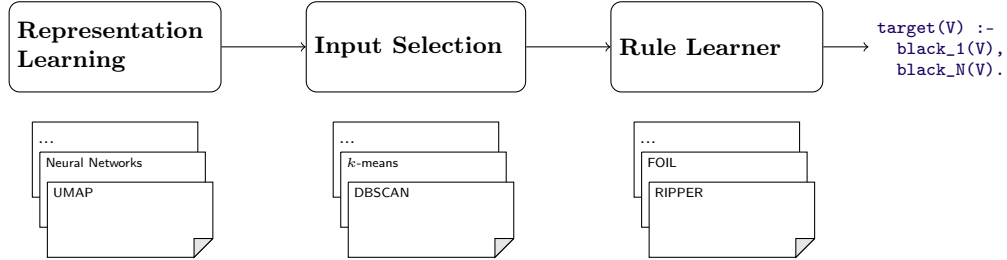
 
There are some related approaches~\cite{nguyen,burkhardt,granmo2019convolutional} considering in particular the MNIST and  Fashion-MNIST data sets as well as similar ones with the main goal of interpretability of the results. 

Nguyen et al.~\cite{nguyen} learn features from given training data by least generalisation. Instead of deterministic rules they rather learn a kind of prototype of each class comparable to a \emph{centroid} generated for instance during k-means clustering. For classification, they basically compare the input image with each learned prototype and predict the one where the sum of differences between pixel values in each location is minimal. This procedure is similar to what is done during clustering. On the MNIST data set they obtain a precision of 66\% and a recall of 80\%, where we achieve approximately 98\% in both metrics. On the Fashion-MNIST data set they obtain 61\% and 63 \% in precision and recall compared with up to 96\% achieved by our approach.

Burkhardt et al.~\cite{burkhardt} deal with the extraction of rules from binary neural networks. On the one hand they seek to find rules explaining the decisions of a trained neural network and evaluate their results with a corresponding \emph{similarity measure}, but on the other hand they also use the original labels of the data sets to learn rules. In the latter approach they achieve about 90\% accuracy on both the MNIST as well as the Fashion-MNIST data set compared with our results of 98\% and up to 95\%, respectively. 

Granmo et al.~\cite{granmo2019convolutional} introduce the \emph{Convolutional Tsetlin Machine} as an interpretable alternative to CNNs. Basically, they learn several patterns which are used like filters in a CNN but which are easier to interpret, although they seem to be not as  self-explaining as our learned rules. So, their approach is not as easily interpretable as ours but compares favourably with simple CNNs achieving an accuracy of 99\% on MNIST and 91\% on Fashion-MNIST.

Moreover, Wang et al.~\cite{Wang2022FOLDSESE} propose the \emph{Fold-SE} algorithm for scalable XAI especially suited for binary classification of large data sets where the largest data set they consider consists of nearly 150 thousand instances. Among others they also investigate the \emph{Heart Disease} and \emph{Diabetes} data set but probably in a slightly adapted form, since they also apply RIPPER for comparison with their method but report much worse results than we obtain in our experiments. Another reason for this inconsistency could be another version/implementation of RIPPER (we used version 0.3.2 of the \emph{wittgenstein} python package). However, they achieve about 75\% accuracy on these two data sets compared to 88-93\% which we obtain with different deciders in our experiments.   

Beyond that, there is a huge variety of rule learning methods and strategies out there as discussed in detail in \cite{modularapproach} but the focus of this paper lies especially on \foil\ and \ripper\ since they have been mainly investigated in our previous work and we aim to directly build on the insights gained therein. Moreover, in our previous work we have shown that other (more modern) ILP tools are not suited for the application on large input data sets.

Furthermore, there are various methods for rule conflict resolution summarised and compared for instance by Lindgren et al.~\cite{Lindgren}. The voting approach introduced in this paper can be thought of as another method of resolving rule conflicts because in case of different predictions made by the rule learners we seek for the advice of a \emph{decider} telling us which rule is better as outlined in the motivational example (cf. Figure~\ref{fig:votingApproach}). To resolve rule conflicts we consult an unexplainable state-of-the-art method eventually yielding a kind of \emph{end-to-end} explainability as explained in more detail in the following sections, as opposed to the methods discussed in \cite{Lindgren} where either different metrics are used to determine an appropriate rule or the whole set of rules is revised in order to obtain a unique prediction. The latter approach is not feasible for our purposes because especially very large data sets are of our interest where a revision of the whole rule set would correspond to very high computational costs. We compared our approach with two of the methods using a certain metric, namely \emph{CN2}~\cite{CN2} and \emph{Naive Bayes Classification} with \emph{Laplace-1 correction}~\cite{naiveBayes}. For this purpose, we investigated the MNIST digits where we especially considered the examples which are handled by \ref{step2} and \ref{step3} of our approach because in \ref{step1} no rule conflict arises since the predictions by the two learners coincide and in \ref{step4} no rule is available. On the resulting 1108 examples our approach is able to give 1054 (95,13\%) correct predictions. On the contrary, \emph{CN2} achieves 63,63\% and \emph{Naive Bayes} is able to predict 68,14\% of the considered examples correctly. For a further comparison with the results obtained by \cite{Lindgren}, we applied our approach also on the \emph{Car Evaluation}  data set~\cite{misc_car_evaluation_19} investigated therein. Since the concrete experimental setups are not described a direct comparison is not straightforward. However, we predict all of the 64 rule conflicts created by our approach correctly, whereas Lindgren reports at most 84,47\% of correctly resolved rule conflicts.

In addition, note that we also consider decision trees in this work especially for comparison because our way of proceeding is not very well suited for tree approaches as explained in detail in Section~\ref{summary}. However, there are approaches (e.g.~\emph{RuleFit}~\cite{RuleFit}, \emph{inTree}~\cite{inTree}) considering in particular trees and random forests aiming for better performance and higher comprehensibility. \emph{RuleFit}, for instance, uses rules extracted from an ensemble of trees additionally to the original features of a data set and learns a sparse linear model on the extended feature set. On the other hand, \emph{inTree} extracts rules from a collection of trees and includes several methods to process the original rules into a compact set of relevant and non-redundant rules. 

Finally, the basic idea of our approach is somewhat similar to what is done in \emph{Meta-Learning}~\cite{meta_learning}. However, they train models on subsets of the whole input data and the final prediction is given by a so-called \emph{arbiter} which is comparable to our \emph{decider} but at the end of the day it is another model that is trained on a reasonably chosen subset of the input data. So, they rather aim to decrease computational costs and achieve a speed up while accepting slightly poorer accuracies, which is all in all similar to the outcome of our previous work.

\section{Methodology}
\label{Approach}

The voting approach introduced in the following bases on two or more rule learning methods which are trained in a first step resulting in independent sets of rules, where each rule set can be used individually for classification. We mainly focus on the application of \foil\ and \ripper\ in this context. 
Due to its simplicity \foil\ is especially suited for problems where primitive \emph{black/white rules} are learnt (cf. MNIST/Fashion-MNIST in Section~\ref{Experiments}), while the increased complexity of \ripper\ is beneficial for data sets including continuous attributes as, for instance, our case study on dental bills. As a result, the combination of these two methods increases the versatility of the introduced voting approach.

Before we present the basic ideas behind the voting approach introduced in this paper, we briefly describe how we use the above-mentioned rule learning methods in order to solve multi-class classification problems.

\subsection{Multi-class Classification with Rules}

Rule learning methods such as \foil\ or \ripper, for instance, usually yield rules for binary classification. For instance, Figure~\ref{fig:partialRules} shows a possible rule learned on the MNIST digits data set deciding whether a given image depicts a zero or not. However, at the end of the day, our goal is to perform a multi-class classification where we predict the depicted digit on a given input image by combining the binary classifiers trained on each possible label.  In order to do that, we construct a binary vector of length equal to the number of labels. Each entry represents the prediction of a rule set learned in advance to classify whether the corresponding label is consistent with the given input. For instance, in the case of the MNIST digits assume that an image depicting a zero is given to the multi-class classifier. The optimal output for this example is the vector
$(1,0,0,0,0,0,0,0,0,0),$
where only the first set of rules learned especially for the label \emph{zero} predicts \emph{True} and all the other rule sets corresponding to the remaining digits predict \emph{False}.
It might occur that two or more binary classifiers return \emph{True} resulting in an ambiguous multi-class classification. In this case, we additionally take the length of the learned rules, e.g. the number of considered pixels in the rules shown in Figure~\ref{fig:partialRules}, into account and return the label corresponding to the longest satisfied rule. On the other hand, if no binary classifier returns \emph{True}, we consider the percentage of satisfied literals in the body of a rule, e.g. how many of the pixels corresponding to a rule have the intensity as requested by the rule, and predict the label with highest percentage.

Consider for instance Figure~\ref{fig:partialRules}, where a chosen example from the MNIST digits data set is shown together with the following learned rule, denoted (as above) in \textsf{Prolog} syntax:
\begin{lstlisting}[style=prolog]
target(V) :- black_130(V), black_233(V), white_354(V), 
						 black_386(V), white_461(V), black_568(V),
						 black_634(V), black_655(V).
\end{lstlisting}

\begin{figure}[t]
\caption{Illustration of a learned rule in the context of classifying whether a given digit is equal to zero or not. The red crosses correspond to an expected positive pixel intensity (\emph{black\_n(V)}), while the blue diamonds in the middle demand zero pixel intensity (\emph{white\_n(V)}).}
\label{fig:partialRules}  
\begin{center}
  \setlength{\fboxsep}{0pt}
  \fbox{
    \includegraphics[trim=50 55 50 40,clip,width = 0.45\textwidth]{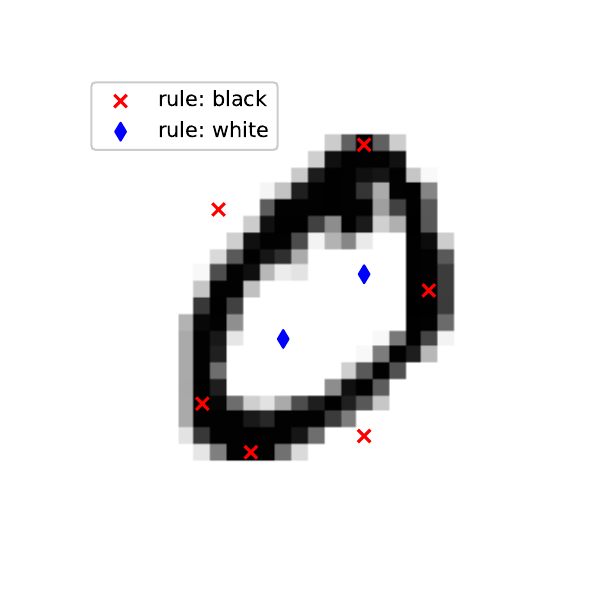}
  }
\end{center}
\end{figure}

Here, the condition \emph{black\_i(V)} expresses that pixel $i$ of the vectorised input image $V$ has positive pixel intensity (cf. Section~\ref{Experiments}), whereas \emph{white\_j(V)} indicates that pixel $j$ of image $V$ is equal to zero. The rule depicted in Figure~\ref{fig:partialRules} consists of 8 conditions in total and obviously 6 of them are fulfilled corresponding to 75\%. Analogously, the percentage of fulfilled conditions is computed for each generated rule and the label corresponding to the highest percentage is predicted. 
The idea behind this procedure is inspired by \emph{prototype generation} (\cite{nguyen}) where during training prototypes of each class are produced and for classification the difference of the input example and the learned prototypes is considered. Likewise, we compute the number of satisfied conditions as a measure of similarity. Moreover, this way of proceeding is basically equivalent to \emph{partial matching} which is a commonly applied technique in the field of rule-based classification (cf.~\cite{partial_matching}). 
In case of ambiguous multi-class predictions we currently weight the rules according to their length. In future work, we want to improve this procedure by introducing a \emph{value of confidence} for the learned rules by, for instance, analysing how often each rule applies in the training set and the corresponding prediction is correct. For example, assume that a learned rule is completely satisfied for 100 training examples but the corresponding prediction is only correct for 90 of these examples. This rule is then equipped with a \emph{value of confidence} of 0,9. 

Note that we prefer this kind of multi-class classification over common approaches like \emph{decision lists}~\cite{decisionlists}, which are basically ordered sets of rules that return a default prediction in case no learned rule is satisfied. Extending the example from Section~\ref{RuleLearning}, we can think of the following result.

\begin{lstlisting}[style=Prolog]
		IF	Sex = Male
		  AND	Has Children = Yes
		THEN	Car = Family
		
		ELSE IF Sex = Male
		  AND	Has Children = No
		Then 	Car = Sport
		
		ELSE ...

\end{lstlisting}

If no rule from the list is satisfied, a default class (usually the largest one) is predicted. For instance, there is an implementation of \ripper\ for multi-class classification~\cite{RipMC} that proceeds in a similar way with the exception that the learned rules are not ordered. This is often also referred to as \emph{decision sets}. This behaviour is the reason why we do not consider this family of rule learners because we aim for a comprehensible rule justifying our prediction instead of returning a default class when no rule is fulfilled. In particular in the context of our use case on dental bills such a way of proceeding would be rather useless because the contained classes are highly imbalanced and there is especially one class representing a high majority of the underlying data.

\subsection{Explainable Voting Approach}
\label{XVoting}

In order to improve the performance of rule learning methods we combine them with (unexplainable) state-of-the-art tools using the ensemble method illustrated in Algorithm~\ref{alg:algorithm} and described in the following.

\begin{enumerate}[label=\textbf{Step \arabic* :}, ref=Step~\arabic*]

\item \label{step1} Our approach considers only the chosen rule learners at first and applies hard voting in order to make a prediction. If, for instance, the predictions from \foil\ and \ripper\ coincide, then we directly return the corresponding class. Note that in this case both \foil\ and \ripper\ provide a rule justifying their prediction. 

\item \label{step2} On the other hand, if the considered rule learners do not yield the same prediction or at least one of them is not able to give a fully justified prediction, we take an additional method into account which serves as \emph{decider}. This means that the voting approach consults the prediction of the decider and checks if one of the chosen rule learners returns the same class. Basically each ML tool is suited as decider but usually it makes the most sense to apply the existing state-of-the-art method at this crucial step. This procedure makes it possible that even if the applied decider is unexplainable, the predicted class is justified by the corresponding rule from the matching learner.  

\item \label{step3} If the second step is also not successful and there is no rule confirming the prediction given by the decider, we investigate the rule sets in more detail and compute the number of satisfied conditions as described above. 
This knowledge makes it possible to define a certain threshold (for instance 70\% as applied in the experiments conducted in Section~\ref{Experiments}) and allow the voting approach to return the prediction given by the decider, if there is at least one corresponding rule where the percentage of satisfied conditions exceeds the chosen threshold. In this case the resulting class is at least partially justified by a deterministic rule and we know exactly which attributes vary from the values demanded by the rule. 

Note that the rule learner is often not able to give a justified prediction (cf. \ref{step2} above) because there is no completely satisfied rule at all. As already mentioned, for direct multi-class classification with rules we consider the number of satisfied conditions in this case and return the label corresponding to the highest percentage. If more than one label achieves this percentage, we additionally consider the length of the satisfied rule (i.e. the number of conditions) and choose the longest (partially) satisfied rule. If this prediction is still ambiguous, the rule learner is not able to give a prediction.
\ref{step3} of our voting approach circumvents this scenario and chooses the prediction from the decider instead, as long as there exists a corresponding rule with a satisfying percentage of fulfilled conditions higher than the chosen threshold.

Note that in this case we always additionally check whether the addressed percentage equals the maximal probability returned by the corresponding learner.
To understand the necessity of this additional check think for instance of a binary classification problem where a rule learner returns a probability of 80\% for label 0 and at the same time a probability of 75\% for label 1. In this edge case we could ``justify'' each prediction of the decider with the chosen threshold of 70\% which contradicts the intention behind our approach. 
Considering multi-class classification we apply an additional \emph{tolerance} value (which is equal to 10\% in the experiments below) and check whether the addressed percentage is higher than the maximal returned probability minus this tolerance value. For instance, we could obtain a probability of 80\% for the digit five and at the same time 75\% for the digit six, while the decider predicts a six. The \emph{tolerance} allows us to incorporate the prediction from the decider, although the corresponding probability is slightly smaller than the maximal probability. This addition makes especially sense for data sets like the MNIST images where several labels are possible and a small change of the intensity of one pixel (eg. due to thresholding) can change the whole prediction.

\item  \label{step4} Otherwise, if none of the above cases applies, we say that our approach is not able to give an explainable prediction since we cannot convincingly justify the prediction from the decider.\\

\end{enumerate}

\begin{algorithm}[b!]
    \caption{Pseudo-Code for Voting Approach}
    \label{alg:algorithm}
    \textbf{Input}: Predictions from rule learners and decider\\
    \textbf{Parameter}: \emph{Threshold} and \emph{Tolerance} between 0 and 1\\
    \textbf{Output}: Final explainable prediction
    \begin{algorithmic} 
        \IF {rule learners coincide}
        \STATE \textbf{return} rule learner prediction
        \ELSIF {a rule learner coincides with decider}
        \STATE \textbf{return} prediction according to the rule chosen by the decider
        \ELSIF {threshold is exceeded \AND percentage $\geq$ MAX - tolerance}
        \STATE \textbf{return} prediction according to the rule chosen by the decider
        \ELSE
        \STATE \textbf{return} no prediction
        \ENDIF
    \end{algorithmic}
\end{algorithm}

In summary, we generate sets of rules and fit a state-of-the-art decider method at first. Afterwards, we apply the trained methods for classification and, finally, we combine the obtained predictions using the ideas introduced above and outlined in Algorithm~\ref{alg:algorithm}. A concrete example of the procedure illustrating both coinciding as well as different predictions of the rule learners is shown in the motivational example in the introduction (cf.~Figure~\ref{fig:votingApproach}).

\pagebreak

\section{Experimental Evaluation}
\label{Experiments}

In this section we evaluate the voting approach introduced above on several common benchmark data sets which are also investigated in existing (unexplainable) ensemble approaches as well as a practical example from insurance business. 

\subsection{Experimental Setup}

On the data sets explained in the following we always apply the rule learning methods \foil\ and \ripper\ as explained in more detail in the Appendix in Section~\ref{Techniques} in a first step resulting in two independent sets of rules. Note that whenever no train/test split is given, we randomly extract a training set consisting of 80\% of the given instances while the remaining 20\% are used as test data. So, the rules learned on the training set are afterwards evaluated on the test set. For the smaller benchmark datasets (except for MNIST and Fashion-MNIST) we repeat the experiments 10 times and show the mean accuracy in Table~\ref{accuracies} and Table~\ref{tab:cn2}. In the same manner an unexplainable decider method is trained and evaluated as described in more detail in Section~\ref{summary}. 

This is where our voting approach comes into play. Given the predictions from \foil\ and \ripper\ as well as from the decider we proceed as described in Section~\ref{XVoting}. Note that we choose the same parameters for all the considered data sets, namely a \emph{threshold} of 0.7 and, for the multi-class problems, a \emph{tolerance} of 0.1. Recall that both of these values are used during \ref{step3} where no completely satisfied rule could be found to make a prediction beforehand. Nevertheless, it makes sense to investigate the rule sets in more detail as depicted in Figure~\ref{fig:partialRules} because single instances often differ from the set of examples used by the learner to generate a rule (cf. Section~\ref{Techniques}) only in a very small part of the attributes. Since the learned rules are in general very short (i.e. they usually consist of less than 5 conditions as shown below for each considered data set), we choose the threshold to be equal to 0.7. Think, for instance, of a rule consisting of 4 conditions. Choosing a higher threshold of 0.8, for example, \ref{step3} of our approach would never apply and a smaller threshold is not reasonable any more as one can imagine from Figure~\ref{fig:partialRules} where one could draw basically every digit satisfying 50\% of the conditions of the illustrated rule.

Moreover, for the considered multi-class problems (i.e. MNIST and Fashion-MNIST as well as the case study on dental bills) we define the tolerance as 0.1 because of similar considerations as explained above for the threshold. Using a smaller value, \ref{step3} would hardly apply and larger values would result in high ambiguity.

Eventually, our approach yields explainable predictions which are justified by at least 70\% of a learned rule. In future work we consider using the decider to slightly adapt the learned rule sets to prevent cases where the prediction is only partially justified as explained in Section~\ref{Conclusion}.

Before going into detail on the obtained results, we briefly explain the underlying data.

\subsubsection{Spambase.}

This data set from the UCI Machine Learning Repository has been investigated for instance by Jain et al.~\cite{spambase-citation} where they achieve an accuracy of 94.6\% using an unexplainable ensemble approach. The data set consists of 4601 examples with 57 attributes concerning the frequency of certain words appearing in an e-mail in \% as well as the length of uninterrupted sequences of capital letters. More precisely, the considered attributes are defined as follows:\footnote{See \url{https://archive.ics.uci.edu/ml/datasets/spambase}.}
\begin{itemize}
\item 48 continuous real attributes of type \emph{word\_freq\_WORD} in the intervall $[0,100]$ corresponding to the percentage of words in the e-mail that match \emph{WORD}. Examples for the value of \emph{WORD} are for instance \emph{make}, \emph{business} and \emph{money}.
\item  6 continuous real attributes of type \emph{char\_freq\_CHAR} in the intervall $[0,100]$ corresponding to the percentage of characters in the e-mail that match \emph{CHAR}, where \emph{CHAR} $\in$ \{\emph{;,(,[,!,\$,\#}\}.
\item 1 continuous real attribute of type \emph{capital\_run\_length\_average} in the range $[1, \dots]$ corresponding to the average length of uninterrupted sequences of capital letters.
\item 1 continuous integer attribute ($\in \mathbb{N}$) of type \emph{capital\_run\_length\_longest} denoting the length of the longest uninterrupted sequence of capital letters.
\item 1 continuous integer attribute ($\in \mathbb{N}$) of type \emph{capital\_run\_length\_total} containing the total number of capital letters in the e-mail.
\end{itemize}
This is a binary classification problem with the aim to differentiate between spam and non-spam e-mails.
In our experiments we round the attributes concerning percentages (i.e. the first 54 attributes) to the first decimal place and the other attributes to the nearest ten for values in the intervall $[10,100]$, while higher values are rounded to the hundredths place. Note that the attribute \emph{capital\_run\_length\_average} is casted to integer beforehand. We apply this kind of preprocessing in order to make the data set better suitable for the chosen rule learners, in particular \foil\ (cf. Section~\ref{Techniques}).

An example of a rule learned by \ripper\ in order to detect a spam e-mail is the following:

\begin{lstlisting}[style=Prolog]
		IF	word_freq_hp <= 0.5
		  AND	word_freq_remove >= 0.4
		  AND	word_freq_business >= 0.5
		THEN	Type = Spam.

\end{lstlisting}

\subsubsection{Heart Disease.}

This data set is also part of the UCI Machine Learning repository. More precisely, we examine the \emph{Cleveland} data set\footnote{See \url{https://archive.ics.uci.edu/ml/datasets/Heart+Disease}.} consisting of 303 instances with 13 attributes corresponding to some factors that correlate with risks of heart diseases such as blood pressure or cholesterol, for instance (cf. Table~\ref{tab:heartdisease}). An ensemble approach investigating in particular this data set has been published for instance by Abdulsalam et al.~\cite{iasc.2023.032262}, where Shap Values are used trying to explain the obtained model which achieves an accuracy of 90.16\%. However, note that they consider a single train-test-split whereas our documented results are the mean of 10 repetitions. So, for a more representative comparison we include a row in Table~\ref{accuracies} showing our results for the same train-test-split. The other tables always depict the average results on 10 repetitions.
For our experiments, we round the values presenting age, blood pressure, cholesterol and heart rate to the tens place and cast ST depression to integer enabling a more efficient application of the rule learners. Moreover, there are 6 missing values included in this data set which we impute by calculating the mean value of the corresponding attribute. 
An example of a rule learned by \ripper\ in order to detect a probable heart disease is the following:

\begin{lstlisting}[style=Prolog]
		IF	Slope = 2
		  AND	Exang = 1
		  AND	Thal = 7
		THEN	Heart Disease = Yes.

\end{lstlisting}

\begin{table*}
\caption{Information About the Attributes of the \emph{Cleveland} Data Set.}
\smallskip  
\label{tab:heartdisease}
\centering
\begin{tabular*}{\textwidth}{@{\extracolsep{\fill}}llp{45ex}}
\toprule
\textbf{Attribute} & \textbf{Type} & \textbf{Description} \\
\midrule
Age & Continuous ($\in [29,77]$) & Age in years \\
\hline
Sex & Discrete ($\in \{0,1\}$) & 1 = male \\
&&							    0 = female \\
\hline
Cp & Discrete ($\in \{1,2,3,4\}$) & Chest pain type \\
&&1 =  typical angina \\
&&2 =  atypical angina \\
&&3 =  non-anginal pain \\
&&4 =  asymptomatic \\
\hline
Trestbps & Continuous ($\in [94,200]$) & Resting blood pressure in mm/Hg \\
\hline
Chol & Continuous ($\in [126,564]$) & Cholesterol in mg/dl \\
\hline
Fbs & Discrete ($\in \{0,1\}$) & Fasting blood sugar $>$ 120 mg/dl \\
&& 1 = true\\
&& 0 = false\\
\hline 
Restecg & Discrete ($\in \{0,1,2\}$) & Resting electrocardiographic results \\
&& 0 = normal \\
&& 1 = having ST-T wave abnormality\\
&& 2 = probable or definite left ventricular hypertrophy\\
\hline
Thalach & Continuous ($\in [71,202]$) & Maximum heart rate achieved\\
\hline
Exang & Discrete ($\in \{0,1\}$) & Exercise induced angina\\
&& 1 = Yes \\
&& 0 = No \\
\hline
Oldpeak & Continuous ($\in [0,6.2]$) & ST depression induced by exercise relative to rest \\
\hline
Slope & Discrete ($\in \{1,2,3\}$) & Slope of the peak exercise ST segment \\
&& 1 = upsloping\\
&& 2 = flat\\
&& 3 = downsloping\\
\hline
Ca & Discrete ($\in \{0,1,2,3\}$) & Number of major vessels colored by flouroscopy\\
\hline
Thal & Discrete ($\in \{3,6,7\}$) & Heart rate\\
&& 3 = normal\\
&& 6 = fixed defect\\
&& 7 = reversable defect\\
\bottomrule
\end{tabular*}
\end{table*}

\subsubsection{Diabetes.}

The Pima Indian diabetes dataset from \emph{kaggle}\footnote{See \url{https://www.kaggle.com/datasets/uciml/pima-indians-diabetes-database}.} consists of 768 instances with 8 attributes corresponding to factors that influence the risk of having Diabetes Mellitus (cf. Table~\ref{tab:diabetes}).
Similar as for the heart disease data set we preprocess the attributes as follows. For \emph{pregnancies} we group all values higher than 5 in one class. \emph{Glucose}, \emph{BloodPressure}, \emph{SkinThickness}, \emph{Insulin} and \emph{Age} are rounded to the tens place. In addition, for \emph{Insulin} we round values higher than 200 to the hundredths place. \emph{BMI} is transformed to multiples of 5 and we round the \emph{DiabetesPedigreeFunction} to the first decimal place. Moreover, we impute missing data by taking the median of the target value like in \cite{s22197268} where they achieved an accuracy of 90\% on this data set using Shap Values as well as Lime plots to explain their model. However, note that unlike them we do not apply class balancing.  
An example of a rule learned by \ripper\ in order to predict diabetes is the following:

\begin{lstlisting}[style=Prolog]
		IF	BloodPressure in [70,80]
		  AND	Insulin in [140,170]
		THEN	Diabetes = Yes.

\end{lstlisting}

\begin{table*}
\caption{Information About the Attributes of the Pima Indian Diabetes Data Set.}
\label{tab:diabetes}
\smallskip
\centering
\begin{tabular*}{\textwidth}{@{\extracolsep{\fill}}llp{25ex}}
\toprule
\textbf{Attribute} & \textbf{Type} & \textbf{Description} \\
\midrule
Pregnancies & Continuous ($\in [0,17]$) & Number of times pregnant\\
Glucose & Continuous ($\in [44,199]$) & Plasma glucose concentration\\
BloodPressure & Continuous ($\in [24,122]$) & Diastolic blood pressure (mm Hg)\\
SkinThickness & Continuous ($\in [7,99]$) & Triceps skin fold thickness (mm)\\
Insulin & Continuous ($\in [14,846]$) & 2-Hour serum insulin (mu U/ml)\\
BMI & Continuous ($\in [18.2,67.1]$) & Body mass index ($\frac{\text{weight in kg}}{(\text{height in m})^2}$)\\
DiabetesPedigreeFunction & Continuous ($\in [0.078,2.42]$) & Diabetes pedigree function\\
Age & Continuous ($\in [21,81]$) & Age (years)\\
\bottomrule
\end{tabular*}
\end{table*}

\subsubsection{COVID-19~\cite{covid-dataset}.}

This data set contains informations about 1736 patients where in particular certain blood values that might indicate a COVID-19 infection are considered. Analogously as for the above data sets, we round the attributes and impute missing data by taking the mean of the corresponding attribute. Note that for the rounding we consider the amount of different values of each attribute and cast all attributes with more than 10 possible values to integer. Afterwards, all attributes with more than 15 different values are rounded to the tens place and, finally, all values in the data set that are higher than 200 are rounded to the hundredths place. Moreover, we construct a subset consisting of 1350 patients where we remove instances with more than 4 missing values as done by Gong et al.~\cite{10.3389/fpubh.2022.874455} who also investigated this data set and achieved 80.4\% accuracy on this subset, although they report a remaining number of 1374 instances. So, they might have used a slightly different data set. In addition, the features \emph{CK} and \emph{UREA} are dropped because of their high percentage of missing values. 
In our experiments we consider both the original data set as well as the constructed subset which we call \emph{Covid\_restricted} in the following.

An example of a rule learned by \ripper\ in order to predict COVID-19 is the following:

\begin{lstlisting}[style=Prolog]
		IF	ALP in [70,80]
		  AND	GGT in [60,70]
		THEN Covid = Yes,

\end{lstlisting}

where \emph{ALP} denotes the \emph{Alkaline phosphatase} in U/L and \emph{GGT} is short for \emph{Gamma glutamyltransferase} which is also measured in U/L.

\subsubsection{MNIST.}

This data set consists of $28 \times 28$ grayscale images of handwritten digits (cf.~Figure \ref{fig:partialRules}) which are split into 60 thousand training examples and 10 thousand test examples, where all digits are evenly represented. For our experiments we normalise and vectorise the given images. Afterwards, we apply thresholding such that a pixel intensity higher than 0.05 corresponds to a black pixel and a lower value represents a white pixel. So, in summary, binary vectors of length 784 serve as input in this example.

The state-of-the-art accuracy of 99.91\% on this data set is achieved by An et al.~\cite{An2020AnEO} who apply ensembles of convolutional neural networks.

\subsubsection{Fashion-MNIST.}

This data set is very similar to the MNIST digits data set. It consists of $28 \times 28$ grayscale images of pieces of clothing such as shoes, dresses, etc., where we use the same preprocessing as above, i.e. we end up with binary vectors of length 784. The learned rules on this data set are similar to the ones learned on the MNIST digits and demand that certain pixels are black or white, respectively (cf. Figure~\ref{fig:partialRules}). 

The state-of-the-art accuracy of 96.91\% on this data set is achieved by Tanveer et al.~\cite{tanveer21} who apply a fine-tuned differential architecture search (DARTS~\cite{liu2019darts}).\\

Note that the two MNIST data sets are very commonly used benchmarks for various machine learning tools, especially for neural networks~\footnote{See \url{https://paperswithcode.com/sota/image-classification-on-mnist} and \url{https://paperswithcode.com/sota/image-classification-on-fashion-mnist}, respectively.}. We are in particular interested in these two data sets because they are similar to the data considered in the case study presented in Section~\ref{CaseStudy} due to the relatively large number of instances and attributes.

\subsection{Objectives \& Summary}
\label{summary}

The primary objective of the empirical evaluation is to evaluate the here proposed voting approach for explainable classification
with rule learning. In particular, our empirical evaluation sought to answer the following questions.

\newcommand*{\RQ}[1]{\textbf{RQ{#1}}}
\makeatletter 
\let\orgdescriptionlabel\descriptionlabel
\renewcommand*{\descriptionlabel}[1]{%
  \let\orglabel\label
  \let\label\@gobble
  \phantomsection
  \protected@edef\@currentlabel{#1\unskip}%
  \let\label\orglabel
  \orgdescriptionlabel{#1}%
}
\makeatother

\begin{description}
\item[\RQ 1 \label{q1}]\emph{Comparison to the base method.}
  Can the voting approach provide better accuracy (recall, precision etc.)
  of classification prediction than the base method, i.e. the ordinary rule learning method.

\item[\RQ 2 \label{q2}]\emph{Comparison to unexplainable state-of-the-art methods.}
  How much does accuracy (recall, precision etc.) in the voting approach deviate from unexplainable
  methods provided by related works.
  
\item[\RQ 3 \label{q3}] \emph{Industrial case study.}
  Do we obtain (partially) justified predictions that can compete with unexplainable methods for the classification of
  dental bills, an industrial use case.
  
\item[\RQ 4 \label{q4}] \emph{Level of explainability.}
  What is the obtained level of explainability and how does it compare to standard measures
  like Shap values or Lime plots? 
\end{description}

In order to investigate these questions, we consider the above-mentioned data sets. Note that the results shown in Table~\ref{accuracies} are always obtained on the test data. Moreover, keep in mind that our main goal is not to find a best-possible unexplainable method but to enhance the performance of existing rule learning methods by supporting them with black-box tools. Therefore, we will not spend a lot of time on fine-tuning the decider but rather use a nevertheless very performant boosted decision tree created with the Python \emph{xgboost} package or for the MNIST data sets a convolutional neural network with 7 layers. However, to illustrate the theoretical potential of our approach we also use the ground truth as decider denoted with \emph{Voting + Truth} where the original labels represent a hypothetically perfect decider. Of course, such a decider might be highly biased and is not realistic in most cases, but the comparison should illustrate the best possible outcome our approach could achieve and emphasize the fact that improvements in (unexplainable) state-of-the-art methods can go hand in hand with improvements in end-to-end explainable classifications.

For comparison, Table~\ref{accuracies} also shows the results obtained by a decision tree generated with Python using the method \emph{DecisionTreeClassifier} from the package \emph{sklearn.tree}. Although a decision tree also yields comprehensible rules (cf. Section~\ref{RuleLearning}) its application within our voting approach is not very meaningful as explained in the following example.\\

\begin{table*}[b!]
\caption{Test examples (in percent), where more than 3 different labels could be ``justified'' according to the criteria defined within our approach.}
\label{tab:inapplicability}
\smallskip
\centering
\begin{tabular*}{\textwidth}{@{\extracolsep{\fill}}lrrrr}
\toprule
Data  & \foil\ & \ripper\ & Decision Tree\\
\midrule
MNIST & 7.15 & 3.86 & 70.49 \\
Fashion-MNIST & 9.69 & 6.98 &  41.79 \\
\bottomrule
\end{tabular*}
\end{table*}

Due to the structure of a tree there is a unique prediction for \emph{all} given input examples because eventually we always end up in one leave and return the corresponding class. On the contrary, concerning rule learning methods, the classifier consists of several independent rule sets (one for each label), where each set is used as binary classifier. In this case it is possible that more than one or no binary classifier applies resulting in an ambiguous prediction or a more detailed investigation of the number of fulfilled conditions for each rule. Our voting approach is especially intended to resolve such rule conflicts which are not present for decision trees.

Even if the predictions from a tree and the applied decider differ and we want to investigate the tree rules in more detail during \ref{step3} the percentages of fulfilled conditions have hardly any expressiveness for the rules learned by a decision tree because rules for different labels often coincide until the last branch before the leaf resulting in very high percentages for various labels. For instance, in Table~\ref{tab:inapplicability} we consider the multi-class data sets MNIST and Fashion-MNIST and show the number of examples where more than 3 different labels could be ``justified'' according to the criteria stated in \ref{step3} of our approach, i.e. the addressed percentage is greater or equal than the chosen threshold of 0.7 and higher than the maximal percentage minus a chosen tolerance which is equal to 0.1 in our experiments. So, for a decision tree, in particular, this means that we compute the number of labels where the percentage of fulfilled conditions of a corresponding rule exceeds 0.9 because the maximal percentage is always equal to 1 as already mentioned above. In Table~\ref{tab:inapplicability} we can observe that the tree-rules could be used to ``justify'' more than 3 labels for more than 70\% of the MNIST digits and more than 40\% of the Fashion-MNIST examples which obviously contradicts the intention behind our approach because we aim for meaningful rules with a small ambiguity. So, we consider in particular \foil\ and \ripper\ within our voting approach and use decision trees only for comparison.\\

Note that \foil\ and \ripper\ are of special interest to us since we have investigated especially these two methods in previous work~\cite{modularapproach} and we aim to improve the corresponding results by extending the modular approach shown in Figure~\ref{fig:modularapproach} with an ensemble approach. More modern rule learning methods have also been investigated in previous work but are mostly not applicable on vast data sets as shown in \cite{modularapproach}. However, as already mentioned in Section~\ref{Introduction}, the applied rule learning methods can be replaced by any explainable method yielding \emph{if-then-else} rules or analogous outcomes as shown in Table~\ref{tab:cn2} where CN2 is applied as an alternative rule learner on the benchmark data sets.

\begin{figure}[h!]
\caption{Illustration of Accuracies shown in Table~\ref{accuracies}.}
\label{fig:accs}  
\begin{center}
  \setlength{\fboxsep}{0pt}
  \fbox{
    \includegraphics[width = \textwidth]{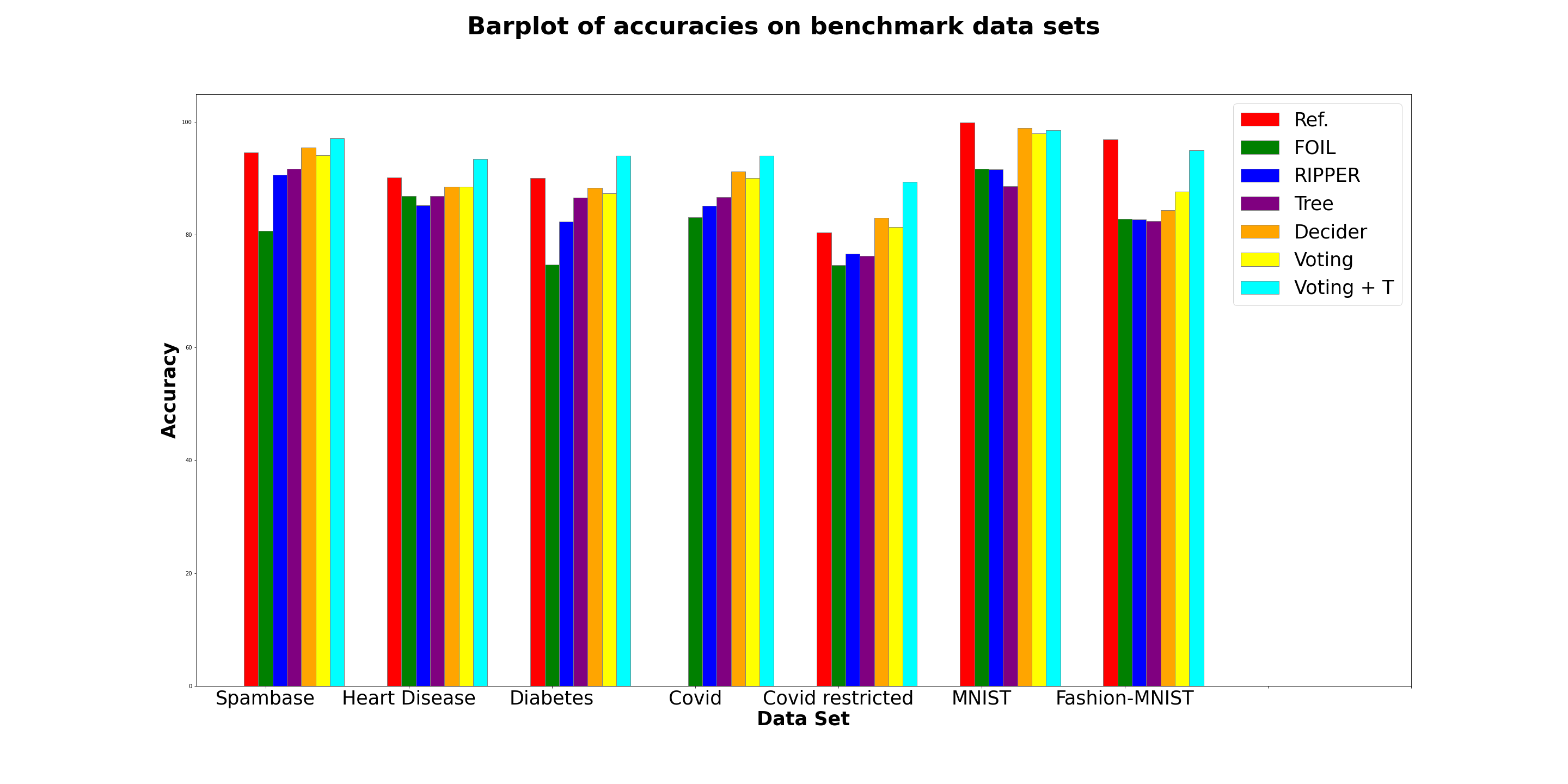}
  }
\end{center}
\end{figure}

The results shown in Table~\ref{accuracies} and Figure~\ref{fig:accs} clearly illustrate the advantages of our approach and, thereby, answer Questions \ref{q1} and~\ref{q2}, respectively, as follows. On all considered benchmarks we achieve better results than the explainable methods alone and outperform them by 2-13\% using a performant but not optimal decider. Very interesting is the fact that on the Fashion-MNIST data set our voting approach yields even a higher accuracy than the applied decider.
Moreover, the potential of our approach is highlighted by the results obtained with the true labels as decider where we achieve more than 90\% accuracy on all benchmarks except for the restricted Covid data set. Note that these results are justified by an easily comprehensible rule and this suggests that an improvement of unexplainable methods goes hand in hand with the improvement of the explainable results obtained by the voting approach introduced in this paper.

\begin{table*}[h!]
    \caption{Comparison of Accuracies (in percent) Obtained by Explainable and Unexplainable Methods. The table compares the accuracies obtained by the
    unexplainable methods in \emph{related works} (column~\emph{Ref.});
    \emph{rule learning} methods (\foil, \ripper\ and decision trees);
    the (unexplainable) \emph{decider} methods (\emph{Decider});
    the proposed voting approach using the decider (\emph{Voting}) and
    finally the voting approach using the true labels (\emph{Voting+T}).
  }
\label{accuracies}
\centering
\begin{tabular*}{\textwidth}{@{\extracolsep{\fill}}lrrrrrrr}
\hline
  \emph{Data}  & \emph{Ref.} & \foil\ & \ripper\ & \emph{Tree} & \emph{Decider} & \emph{Voting} & \emph{Voting+T}
  \\
\hline
  Spambase  & $94.60$  & $80.65$ & $90.56$ & $91.67$ & $95.44$ & $94.07$ & $97.09$
  \\
  Heart Disease & --  & $73.28$  & $72.95$ & $74.92$ & $81.31$ & $77.70$ & $85.25$
  \\
  Heart Dis.(1 split)\footnote{Analogously as done in \cite{iasc.2023.032262}, a single train-test-split with a random state of $42$ is applied.} & $90.16$  & $86.89$  & $85.25$ & $86.89$ & $88.52$ & $88.52$ & $93.44$
  \\
  Diabetes    & $90.00$  & $74.68$  & $82.27$ & $86.56$ & $88.31$ & $87.34$ & $93.96$
  \\
  Covid  & -- &  $83.10$  & $85.06$ & $86.67$ & $91.21$ & $90.06$ & $94.02$
  \\
  Covid\_restricted & $80.40$  & $74.59$ & $76.59$ & $76.19$ & $82.96$ & $81.30$ & $89.33$
  \\
  MNIST  & $99.91$  & $91.68$ & $91.56$ & $88.56$ & $98.95$ & $97.95$ & $98.50$
  \\
  Fashion-MNIST & $96.91$ & $82.81$ & $82.72$ & $82.45$ & $84.32$ & $87.60$ & $94.92$
  \\
\hline
\end{tabular*}
\end{table*}

\begin{table*}[t]
  \caption{Comparison of the accuracies obtained by the
    \emph{rule learning} methods (\foil, \ripper\ and \cntwo) and their combination via voting. Note that the same decider as in Table~\ref{accuracies} is applied.
  }
\label{tab:cn2}
\centering
\begin{tabular*}{\textwidth}{@{\extracolsep{\fill}}lcccccc}
\hline
  \emph{Data}  & \cntwo\ & \foil\ & \ripper\  & \cntwo + \foil\ & \cntwo\ + \ripper\ & \foil\ + \ripper\
  \\
\hline
  Spambase & $92.27$ & $80.65$ & $90.56$ & $94.78$ & $95.33$ & $94.07$
  \\
  Heart Disease & $72.30$ & $73.28$  & $72.95$ & $79.67$ & $78.85$ & $77.70$
  \\
  Diabetes & $80.00$ & $74.68$  & $82.27$ & $84.81$ & $86.75$ & $87.34$
  \\
  Covid  & $85.43$ &  $83.10$  & $85.06$ & $90.06$ & $90.11$ & $90.06$
  \\
  Covid\_restricted & $73.93$ & $74.59$ & $76.59$ & $81.59$ & $82.22$ & $81.30$
  \\
  MNIST & $92.01$  & $91.68$ & $91.56$ & $97.02$ & $96.97$ & $97.95$
  \\
  Fashion-MNIST & $83.19$ & $82.81$ & $82.72$ & $86.70$ & $87.06$ & $87.60$
  \\
\hline
\end{tabular*}
\end{table*}

\begin{table*}[h!]
\caption{Comparison of precision (in percent) obtained by explainable methods and the unexplainable \emph{Decider} method, as well as the proposed voting approach using this method as decider on the one hand as well as the true labels (\emph{Voting+T}) on the other hand.}
\label{tab:precisions}
\smallskip
\centering
\begin{tabular*}{\textwidth}{@{\extracolsep{\fill}}lrrrrrr}
\toprule
  Data  & \foil\ & \ripper\ & \emph{Tree} & \emph{Decider} & \emph{Voting} & \emph{Voting+T}
  \\
\midrule
Spambase  &  84.20 & 91.57 & 91.18 & 95.23 & 95.54 & 99.28  \\
Heart Disease & 81.12 & 80.85 & 75.41 & 81.69 & 81.63 & 90.59   \\
Diabetes  & 86.83 & 84.43 & 85.64 & 87.56 & 88.10 & 96.50        \\
Covid     & 87.27 & 87.18 & 86.59 & 91.19 & 91.27 & 95.74   \\
Covid\_restricted & 79.80 & 78.98 & 75.84 & 82.75 & 82.37 & 92.31     \\
MNIST     &  94.06 & 93.77 & 88.44 & 98.94 & 98.14 & 98.69     \\
Fashion-MNIST &  84.87 & 84.85 & 82.44 & 85.66 & 89.13 & 96.37    \\
\bottomrule
\end{tabular*}
\end{table*}

\begin{figure}[h!]
\caption{Illustration of Precisions shown in Table~\ref{tab:precisions}.}
\label{fig:precs}  
\begin{center}
  \setlength{\fboxsep}{0pt}
  \fbox{
    \includegraphics[width = \textwidth]{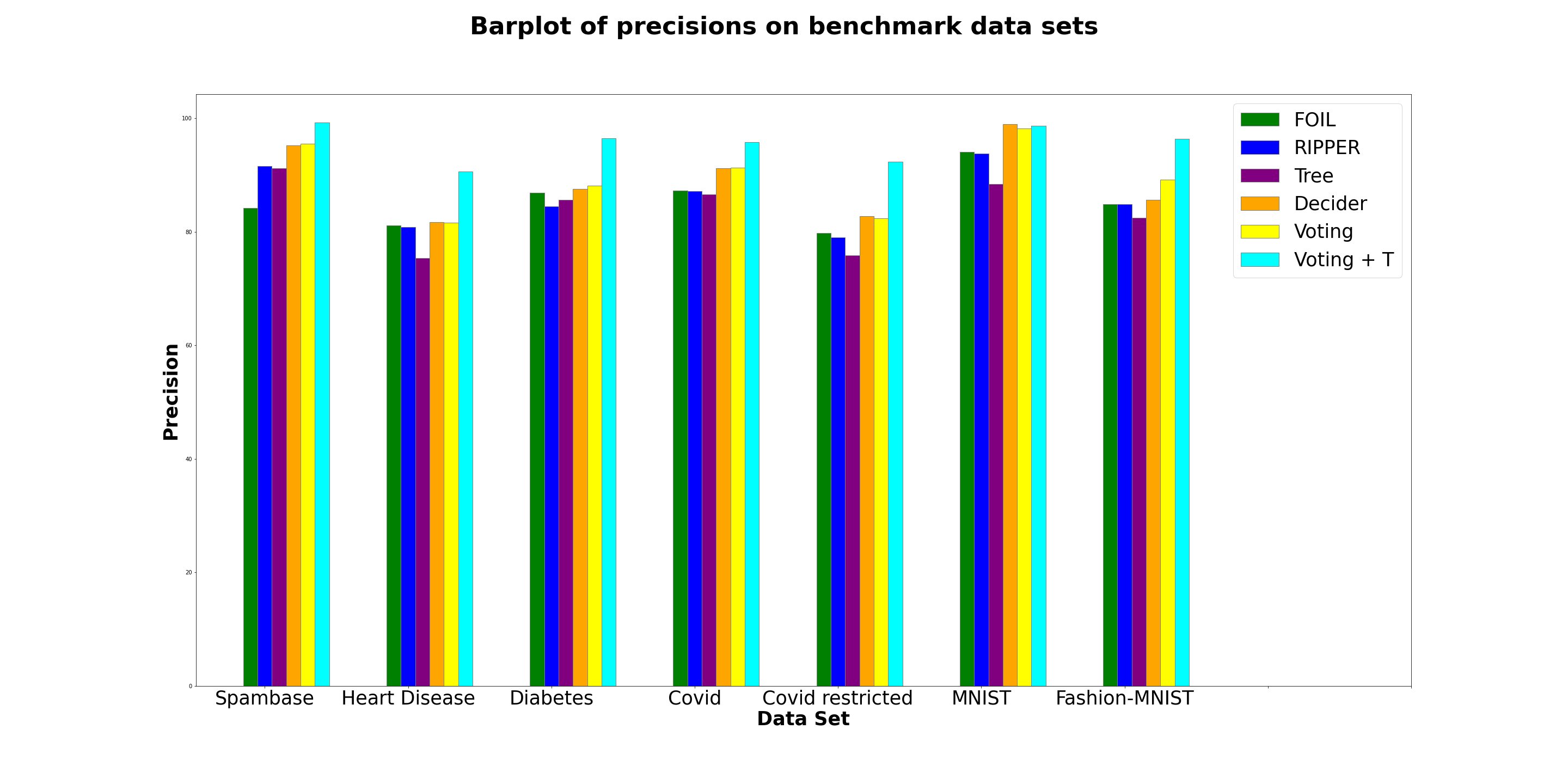}
  }
\end{center}
\end{figure}

\pagebreak
Moreover, especially for our use case on dental bills but also in medicine-related problems such as \emph{Heart Disease} and \emph{Diabetes}, for instance, the \emph{precision} of the results is of utmost importance. Therefore, Table~\ref{tab:precisions} and Figure~\ref{fig:precs} illustrate the resulting precision of our approach compared to the applied methods alone. We can observe that the precision of our voting approach is usually significantly higher than the one achieved with the explainable methods alone. Moreover, it is very remarkable that our voting approach achieves a higher precision than the unexplainable decider method on four out of the seven considered benchmark data sets. On the other examples (i.e. \emph{Heart Disease}, \emph{Covid\_restricted} and \emph{MNIST}) the precision of our voting approach is not even 1\% lower than the one of the unexplainable method.

\begin{table*}[h!]
\caption{Comparison of recalls (in percent) obtained by explainable methods and the unexplainable \emph{Decider} method, as well as the proposed voting approach using this method as decider on the one hand as well as the true labels (\emph{Voting+T}) on the other hand.}
\label{tab:recalls}
\smallskip
\centering
\begin{tabular*}{\textwidth}{@{\extracolsep{\fill}}lrrrrrr}
\toprule
Data  & \foil & \ripper\ & \emph{Tree} & \emph{Decider} & \emph{Voting} & \emph{Voting+T} \\
\midrule
Spambase  &  83.66 & 89.64 & 91.43 & 95.19 & 94.01 & 97.30  \\
Heart Disease & 73.13 & 72.04 & 74.30 & 80.88 & 77.21 & 84.80  \\
Diabetes  &  74.11 & 82.44 & 84.78 & 86.73 & 85.89 & 93.93   \\
Covid     &  83.41 & 85.07 & 86.64 & 91.12 & 90.01 & 94.03     \\
Covid\_restricted  & 73.12 & 75.83 & 75.59 & 82.70 & 80.95 & 88.96  \\
MNIST     & 91.57 & 91.44 & 88.40 & 98.94 & 97.93 & 98.49      \\
Fashion-MNIST & 82.81 & 82.72 & 82.45 & 84.32 & 87.60 & 94.92      \\
\bottomrule
\end{tabular*}
\end{table*}

\begin{figure}[h!]
\caption{Illustration of Recalls shown in Table~\ref{tab:recalls}.}
\label{fig:recs}  
\begin{center}
  \setlength{\fboxsep}{0pt}
  \fbox{
    \includegraphics[width = \textwidth]{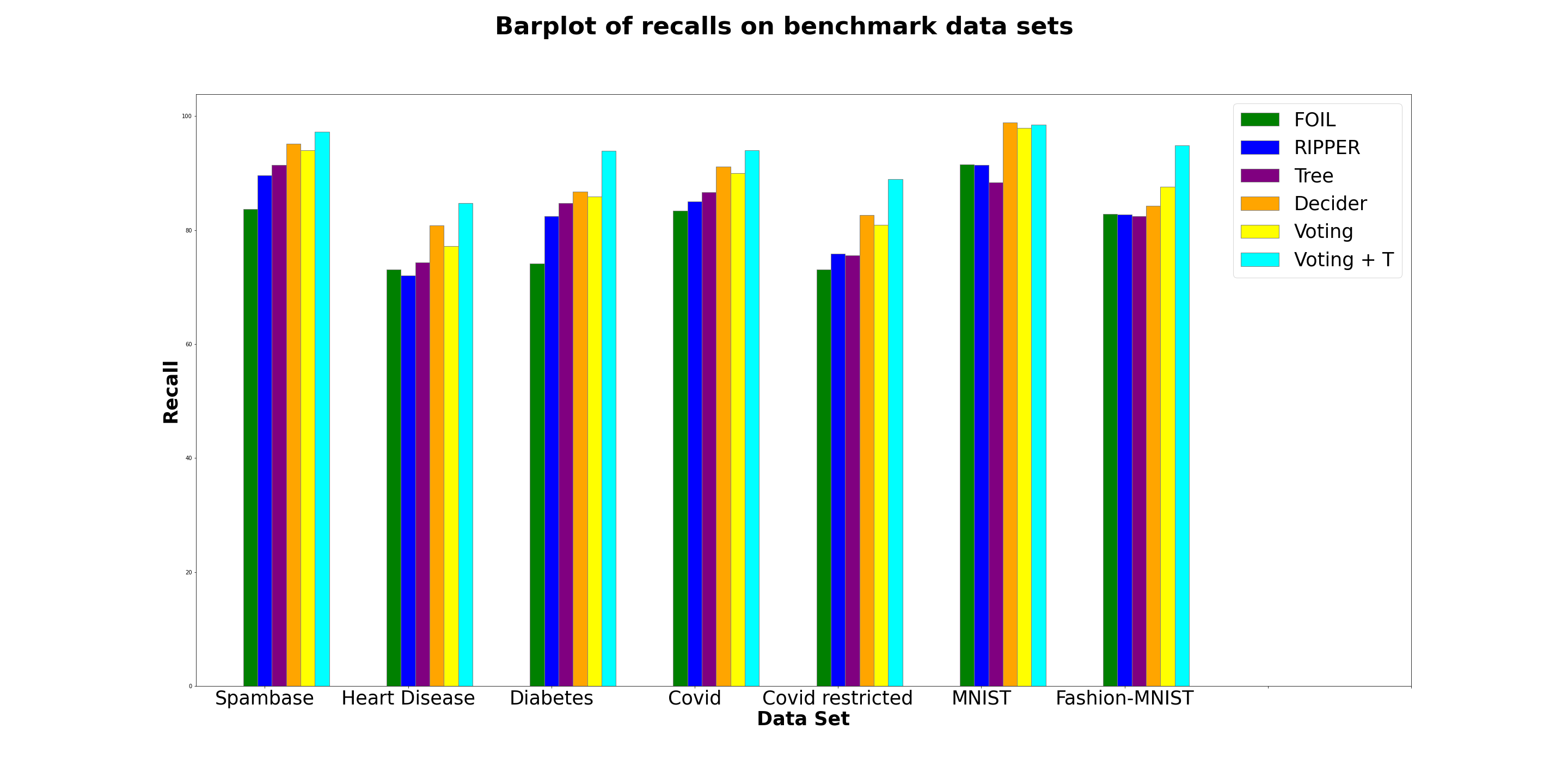}
  }
\end{center}
\end{figure}

\pagebreak
In addition, for sake of completeness, Table~\ref{tab:recalls} and Figure~\ref{fig:recs} show the \emph{recall} obtained by using the explainable voting approach introduced in this paper. Obviously, our approach outperforms the considered explainable methods when it comes to this metric and also compared with the unexplainable decider our approach is better on one of the benchmarks.

Finally, concerning time consumption, we note that the introduced voting approach itself (cf. Algorithm~\ref{alg:algorithm}) produces no significant additional expenses, but it requires the learned rule sets as well as the trained decider method as input. Arguably the total time consumption of the overall procedure is approximately equal to the time needed to fit the above-mentioned methods which, of course, can be done simultaneously. This means that our approach is eventually not much slower than the most time-consuming method applied within the voting approach.

\subsection{Case Study: Dental Bills}
\label{CaseStudy}

The \emph{Allianz Private Krankenversicherung (APKV)} is an insurance company offering health insurance services in Germany. In the course of digitisation, it nowadays becomes a standard to scan medical bills with an OCR engine as well as to offer the possibility to hand in scans of medical bills via, for instance, an app. Those scanned bills are then often automatically processed and the goal is to reimburse the approved sum as fast as possible without a manual processing. The benefits are lower costs and to gain an edge over the competition if the client receives the reimbursed money shortly after handing in the bill, as well. More information about the application of artificial intelligence particularly in the insurance business is given by Lamberton et al.~\cite{dunkelverarbeitung}.

However, a variety of challenges has to be tackled before an automated decision regarding the bill can be made: The quality of the scans may vary, information can be given quite differently and diverse data on the bills can be missing or corrupted. Therefore, imputing missing data is a crucial step in the tool chain.

As decision making, in particular in this sensitive area, should be transparent to both parties, the operational use of black-box machine learning algorithms is often seen critically by the stakeholders and is in many cases avoided. As a consequence, rule learning achieving a comparable performance offers the desired advantage of explainability.  

For our case study, we are focusing on \emph{dental bills}. On those bills, the specific type of dental service per row on the bill is unknown but needed for deciding on the amount of refund. Especially differentiating between material costs and other costs is of crucial importance.

In collaboration with the APKV, we have been provided with a training data set consisting of about half a million instances with 435 attributes.

Using only structured information on the bills such as cost, date and simple engineered features, neural network architectures and tree based gradient boosting have been developed achieving an accuracy of about 90\%. Due to pending non-disclosure agreements we cannot go into detail about the applied methods but we have examined a variety of deep convolutional neural networks as well as \emph{XGBoost}~\cite{xgboost} approaches in order to solve this task using unexplainable methods.\footnote{For more information please directly contact \href{mailto:gabriela.dick_guimaraes@allianz.de}{gabriela.dick\_guimaraes@allianz.de}.} We apply our methodology on these dental bills in order to obtain competitive results in terms of accuracy that in addition lead to a comprehensible classification. Note that the derived rules are of great use, even for non-automated classification of such medical bills to achieve more consistency in the decision making.

\begin{figure}[h!]
\caption{Illustration of results shown in Table~\ref{tab:DentalBills}.}
\label{fig:allianz}  
\begin{center}
  \setlength{\fboxsep}{0pt}
  \fbox{
    \includegraphics[width = \textwidth]{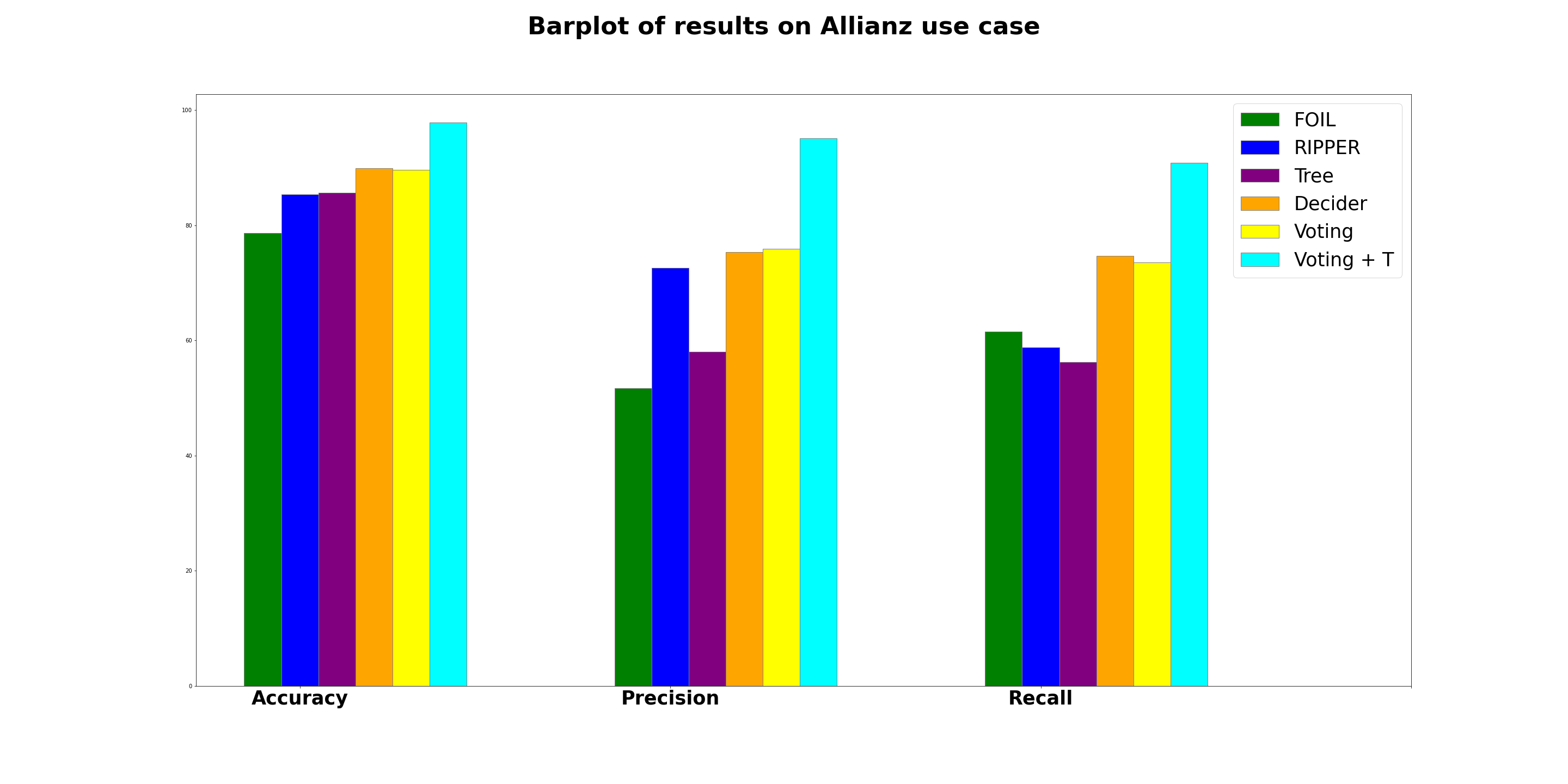}
  }
\end{center}
\end{figure}

Table~\ref{tab:DentalBills} and Figure~\ref{fig:allianz} show that the enhancement of the results obtained by rule learners is not only achieved for some standard benchmark data sets but also for this practical case study. So, also Question~\ref{q3} can be answered positively. As already mentioned, \emph{precision} is a very important metric in insurance business. So, we should especially keep an eye on the corresponding results but note that Table~\ref{tab:DentalBills} illustrates the average precision and recall across the possible classes and the obtained outcomes appear below value due to poor results corresponding to two very small classes.
However, it is remarkable that the voting approach yields even a slightly higher precision than the applied decider. Moreover, analogously as for the standard benchmarks above, we can clearly observe the potential of our approach as we achieve nearly 98\% accuracy with a precision of 95\% using a hypothetically perfect decider. 

\begin{table}[h!]
\caption{Industrial Use Case: Classification of Dental Bills. Results in percent obtained for the case study on dental bills. Note that \emph{Precision} and \emph{Recall} denote the average precision and recall, respectively.}  
\label{tab:DentalBills}
\centering
\begin{tabular*}{\columnwidth}{@{\extracolsep{\fill}}lrrr}
\toprule
\emph{Method}  & \emph{Accuracy} & \emph{Precision} & \emph{Recall} \\
\midrule
  \foil\        & $78.65$     & $51.68$     & $61.54$
  \\
  \ripper\      & $85.31$     & $72.55$     & $58.81$
  \\
  Decision Tree & $85.62$     & $58.05$     & $56.22$
  \\
  Decider (XGB) & $89.90$     & $75.35$     & $74.69$
  \\
  Voting        & $89.65$     & $75.89$     & $73.57$
  \\
  Voting+Truth    & $97.87$     & $95.08$     & $90.80$
  \\
\bottomrule
\end{tabular*}
\end{table}

\subsection{Detailed Analysis} \label{DetailedAnalysis}

As mentioned in Section~\ref{RelatedWork}, explainability approaches in related work such as SHAP and LIME only provide a local approximation for a contribution of an input for a single sample depending on the specified neighbourhood. Therefore, such approaches can not be generalised to the entire machine learning model. Moreover, they are post-hoc approaches.

Contrarily, in this paper we consider easily comprehensible rules. Above, we have already shown that the introduced voting approach significantly enhances the performance of existing rule learning methods answering Questions~\ref{q1}-- \ref{q3}. In the following we will analyse the obtained results in more detail concerning the level of explainability (cf. Question~\ref{q4}).

\begin{table}[t]
\caption{Level of explainability. Overview of the distribution of the predictions in percent on the 3 following levels of explainability: \emph{Explainable} (\ref{step1} and \ref{step2}), \emph{Partially Explainable} (\ref{step3}), \emph{Not Explainable} (\ref{step4})}  
\label{tab:Explainability}
\centering
\begin{tabular*}{\columnwidth}{@{\extracolsep{\fill}}lrrr}
\toprule
\emph{Data}  & \emph{Explainable} & \emph{Part. Explainable} & \emph{Not Explainable} \\
\midrule
  Spambase  & $88.38$ & $9.88$ & $1.74$ 
  \\
  Heart Disease & $88.53$ & $8.20$  & $3.28$ 
  \\
  Diabetes     & $86.36$ & $11.69$  & $1.95$ 
  \\
  Covid        & $91.09$  & $8.05$ & $0.86$ 
  \\
  Covid\_restricted & $83.33$ & $10.37$ & $6.30$ 
  \\
  MNIST     & $96.35$ & $3.46$ & $0.19$ 
  \\
  Fashion-MNIST & $87.00$ & $11.47$ & $1.53$ 
  \\
  Dental Bills & $77.08$ & $21.88$ & $1.05$ 
  \\
\bottomrule
\end{tabular*}
\end{table}

\begin{figure}[h!]
\caption{Illustration of distributions of explainability levels shown in Table~\ref{tab:Explainability}.}
\label{fig:explainability}  
\begin{center}
  \setlength{\fboxsep}{0pt}
  \fbox{
    \includegraphics[width = \textwidth]{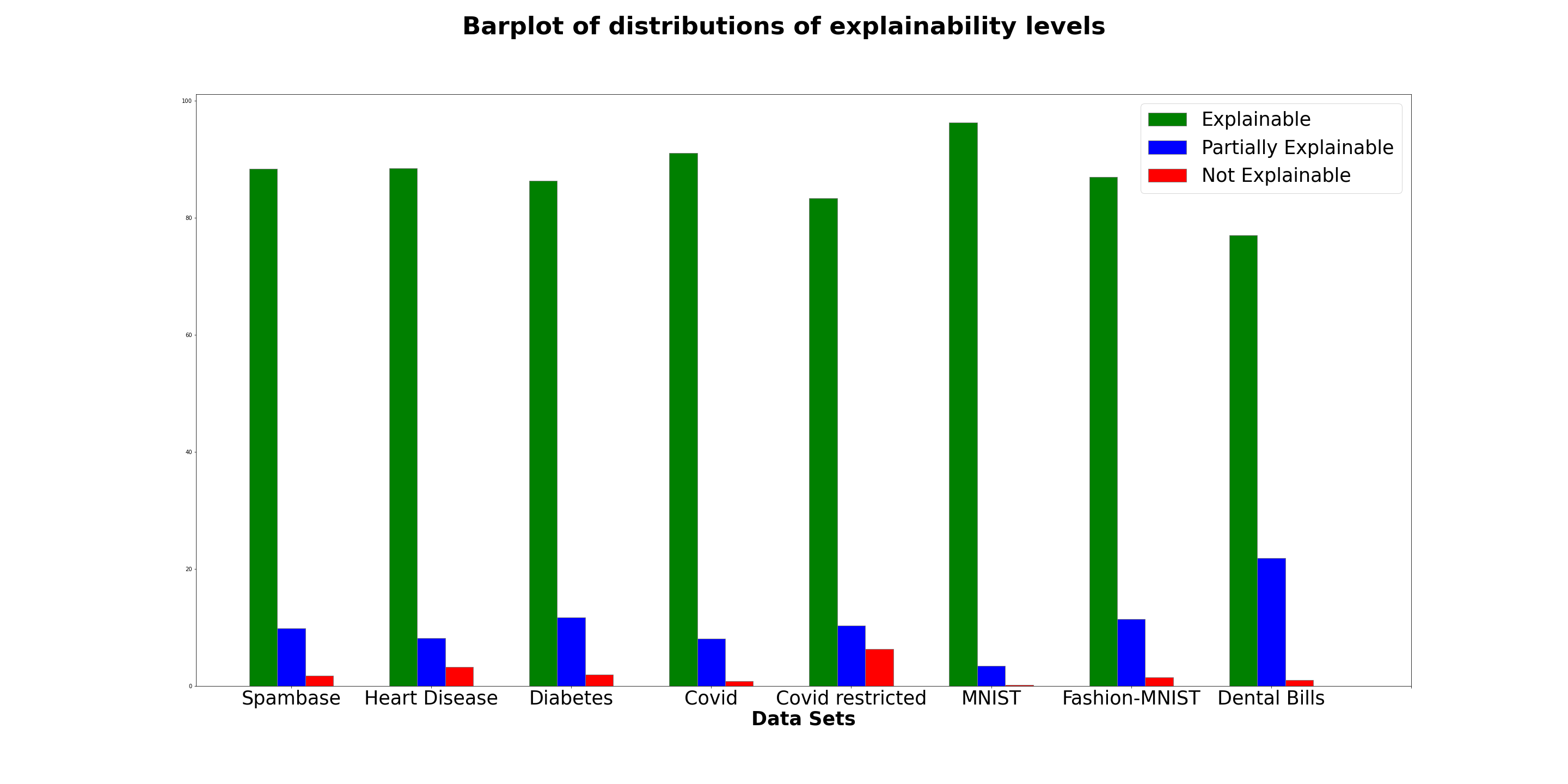}
  }
\end{center}
\end{figure}

Summarised, \ref{step1} and \ref{step2} of our approach yield predictions which are justified by a completely satisfied rule. Predictions returned during \ref{step3} of our approach are at least partially justified fulfilling a large proportion of the conditions of a learned rule, whereas \ref{step4} does not produce any predictions due to the unsatisfactory level of explainability. Table~\ref{tab:Explainability} and Figure~\ref{fig:explainability} show the distribution of the predictions on the above-mentioned levels of explainability. On all standard benchmark data sets nearly 90\% of the given predictions are justified by a completely satisfied rule and also considering the industrial use case on dental bills more than 77\% of the predictions fall into this category. Moreover, it is remarkable that hardly any predictions are not explainable at all. The only outlier is the \emph{Covid\_restricted} data set where about 6\% of the instances cannot be classified using our voting approach. 

Considering in particular the industrial use case on dental bills also the approximately 22\% of partially explainable predictions are of great use as already mentioned in Section~\ref{CaseStudy}. They can be helpful even for non-automated classification because they give a deeper insight into the available data resulting in more consistency in the decision making. At the end of the day, our approach generates rules describing the data. Either the two considered rule learning methods coincide and yield a general description of a class or they differ and point out attributes which might be typical for two classes. The latter case directly identifies examples which might be interesting for closer examination by an insurance specialist, for example to systematically find inconsistencies in the training data which can then later be discussed with experts in health care in order to improve data quality.

\section{Conclusion \& Future Work}
\label{Conclusion}

In this paper, we present a novel kind of voting approach that combines the superior performance of unexplainable black-box methods with the comprehensibility and interpretability of common rule learning methods such as \foil\ and \ripper, for instance, resulting in a very performant but nevertheless (end-to-end) explainable methodology significantly improving the performance of the modular approach introduced in our earlier work.
We show the viability and the achieved benefits of the
introduced methodology in the context of various common benchmark data sets as well as an industrial case study.

Future work will be concerned with the adaption of learned rules by incorporating the prediction given by an unexplainable method. For instance, we consider to apply a kind of optimisation phase similar to the way \ripper\ proceeds where we use the black-box prediction to revise rules with a high percentage of satisfied conditions.
Moreover, we aim to incorporate a \emph{value of confidence} for each rule as a metric of reliability of a rule as an additional way of rule conflict resolution. 

\bibliographystyle{splncs04}
\bibliography{technical_report}













\newpage
\appendix

\section{Rule Learning Algorithms}
\label{Techniques}

Above, we have introduced a novel voting approach combining rule learning methods with (unexplainable) machine learning tools and illustrated the achieved advantages, in particular the gained increase of accuracy. 
In the following, we revise the main ideas behind the rule learners successfully applied on the data sets of interest, i.e. the \foil\ algorithm as well as the \ripper\ algorithm. In addition to self-certainty, this serves to compare the algorithms in more detail as well as to give some insights in our implementation of the \foil\ algorithm.

\subsubsection{FOIL~\cite{foil}.}

\emph{First Order Inductive Learner} (\foil) is one of the first ILP systems and despite its simplicity, it performs quite remarkably on our experiments yielding results on a par with the state-of-the-art \ripper\ algorithm at least for some of the considered benchmarks, for instance MNIST and Fashion-MNIST where primitive \emph{black or white rules} are learned (cf. Table~\ref{accuracies}). The basic procedure, given a set of positive examples $\mathcal{P}$ and a set of negative examples $\mathcal{N}$, is outlined in the upper half of Figure~\ref{fig:foil_ripper}.
Note that the set of possible literals depends on the specific input data and consists of conditions of the form $a = v$ for all attributes $a$ that are not yet included in the considered \emph{NewRule} and all values $v$ occurring in the input data. Moreover, within the \foil\ algorithm a specific \emph{gain function} is applied in order to find the best possible literal. Basically, this means that a literal should cover as much positive examples as possible and at the same time as few negative ones as possible. The performance of the \foil\ algorithm is in particular dependent on the chosen gain function (see \cite{Jimnez2011OnIF}).

For the experiments conducted in Section~\ref{Experiments}, we implemented the \foil\ algorithm on our own because we wanted a running Python implementation
on the one hand and on the other hand our implementation simplifies the application of the algorithm to large, array-like data sets as in particular in the considered case study.%
\footnote{The corresponding code is available at~\url{https://github.com/AlbertN7/ModularRuleLearning}.}

\subsubsection{RIPPER~\cite{ripper}.}
\emph{Repeated Incremental Pruning to Produce Error Reduction} (\ripper) is a rule learning method which is able to handle large noisy data sets effectively. Similar as its predecessor \emph{Incremental Reduced Error Pruning} (\irep~\cite{Frnkranz1994IncrementalRE}), it combines the separate-and-conquer technique applied in \foil\ with the  reduced error pruning strategy~\cite{Brunk1991AnIO}. The pseudocode for the algorithm is outlined in the lower half of Figure~\ref{fig:foil_ripper}, where $\mathcal{P}$ and $\mathcal{N}$ again denote the positive and negative examples, respectively.

Note that the \emph{description length} is a measure of total complexity (in bits) preventing overfitting. It consists of the complexity of the model plus that of the examples which are not captured. As the learned set of rules grows, the complexity of the model increases and the number of uncovered examples decreases. In sum one can say that description length helps to balance between minimisation of classification error and minimisation of model complexity.
The \emph{growing set} typically consists of two third of the input data, whereas the \emph{pruning set} contains the remaining third. As the names already suggest, the growing set is used during the growing phase, where conditions are iteratively added to a rule similar as in the \foil\ algorithm until no negative examples are covered. On the other hand, the pruning set is used during the pruning phase, where the learned conditions are deleted in reversed order to find the pruned rule which maximizes $\frac{p - n}{p + n}$. Here, $p$ and $n$ denote the number of positive and negative pruning examples that are covered by the pruned rule. 

One crucial difference of the \ripper\ algorithm opposed to its predecessor, the \irep\ algorithm, is the additional optimisation phase, which is applied for $k$ iterations after the generation of a rule set. Typically $k$ is equal to 2. During the optimisation each rule is considered in the order in which they have been learned and two alternative rules are constructed for each original rule. The \emph{replacement rule} is learned by growing and pruning from scratch, whereas the \emph{revision rule} is grown by greedily adding conditions to the considered original rule rather than the empty rule. Afterwards, a description length heuristic is used to choose the better one.
In our experiments we made use of the \emph{wittgenstein} package implemented in Python.

\begin{figure}[t]
\caption{Pseudocode Describing the \foil\ and the \ripper\ Algorithm.}
\label{fig:foil_ripper}
\begin{lstlisting}
@(\textbf{FOIL(}$\mathcal{P}, \mathcal{N}$\textbf{)})@
LearnedRules <- {}
while (@($\mathcal{P}$)@ not empty):
	//@(\textbf{Learn a new rule})@
	NewRule <- Rule without any conditions
	CoveredNegs <- @($\mathcal{N}$)@
	while (CoveredNegs not empty):
		//@(\textbf{Add a literal to NewRule})@
		PossibleLiterals <- Set of all possible literals 
		BestLiteral <- @($\argmax_{l~\in~\text{PossibleLiterals}}$ \textbf{Gain(}$l$\textbf{)})@
		NewRule.append(BestLiteral)
		CoveredNegs <- Subset of CoveredNegs satisfying NewRule
	LearnedRules.append(NewRule)
	@($\mathcal{P}$)@ <- Subset of @($\mathcal{P}$)@ not covered by LearnedRules
return LearnedRules
\end{lstlisting}
\begin{lstlisting}
@(\textbf{GenerateRuleSet(}$\mathcal{P}, \mathcal{N}$\textbf{)})@
LearnedRules <- {}
DL <- DescriptionLength(LearnedRules,@($\mathcal{P}, \mathcal{N}$)@)
while (@($\mathcal{P}$)@ not empty):
	//@(\textbf{Grow and prune a new rule})@
	split (@($\mathcal{P}, \mathcal{N}$)@) into (@($\mathcal{P}_{\text{grow}}, \mathcal{N}_{\text{grow}}$)@) and (@($\mathcal{P}_{\text{prune}}, \mathcal{N}_{\text{prune}}$)@)
	NewRule <- GrowRule(@($\mathcal{P}_{\text{grow}}, \mathcal{N}_{\text{grow}}$)@)
	NewRule <- PruneRule(NewRule,@($\mathcal{P}_{\text{prune}}, \mathcal{N}_{\text{prune}}$)@)
	LearnedRules.append(NewRule)
	if (DescriptionLength(LearnedRules,@($\mathcal{P}, \mathcal{N}$)@) > DL + 64):
		 //@(\textbf{Prune the whole rule set and exit})@
		 for each rule @($\mathcal{R}$)@ in LearnedRules in reversed order:
		 	if (DescriptionLength(LearnedRules\{@($\mathcal{R}$)@},@($\mathcal{P}, \mathcal{N}$)@) < DL):
		 		LearnedRules.delete(@($\mathcal{R}$)@)
		 		DL <- DescriptionLength(LearnedRules,@($\mathcal{P}, \mathcal{N}$)@)
		return LearnedRules
	DL <- DescriptionLength(LearnedRules,@($\mathcal{P}, \mathcal{N}$)@)
	@($\mathcal{P}$)@ <- Subset of @($\mathcal{P}$)@ not covered by LearnedRules
	@($\mathcal{N}$)@ <- Subset of @($\mathcal{N}$)@ not covered by LearnedRules
return LearnedRules
		
@(\textbf{OptimizeRuleSet(}LearnedRules,$\mathcal{P}, \mathcal{N}$\textbf{)})@ 
for each rule @($\mathcal{R}$)@ in LearnedRules:
	LearnedRules.delete(@($\mathcal{R}$)@)
	@($\mathcal{P_{\text{unc}}}$)@ <- Subset of @($\mathcal{P}$)@ not covered by LearnedRules
	@($\mathcal{N_{\text{unc}}}$)@ <- Subset of @($\mathcal{N}$)@ not covered by LearnedRules
	split (@($\mathcal{P_{\text{unc}}}, \mathcal{N_{\text{unc}}}$)@) into (@($\mathcal{P}_{\text{grow}}, \mathcal{N}_{\text{grow}}$)@) and (@($\mathcal{P}_{\text{prune}}, \mathcal{N}_{\text{prune}}$)@)
	ReplRule <- GrowRule(@($\mathcal{P}_{\text{grow}}, \mathcal{N}_{\text{grow}}$)@)
	ReplRule <- PruneRule(ReplRule,@($\mathcal{P}_{\text{prune}}, \mathcal{N}_{\text{prune}}$)@)
	RevRule <- GrowRule(@($\mathcal{P}_{\text{grow}}, \mathcal{N}_{\text{grow}},\mathcal{R}$)@)
	RevRule <- PruneRule(RevRule,@($\mathcal{P}_{\text{prune}}, \mathcal{N}_{\text{prune}}$)@)
	OptimizedRule <- better of ReplRule and RevRule
	LearnedRules.append(OptimizedRule)	
return LearnedRules
		 	
@(\textbf{RIPPER(} $\mathcal{P}, \mathcal{N}, k$\textbf{)})@
LearnedRules <- @(\textbf{GenerateRuleSet(}$\mathcal{P}, \mathcal{N}$\textbf{)})@
repeat k times:
	LearnedRules <- @(\textbf{OptimizeRuleSet(}LearnedRules,$\mathcal{P}, \mathcal{N}$\textbf{)})@ 
return LearnedRules
\end{lstlisting}
\end{figure}

\end{document}